%% file: MMP_CVAE.tex
\documentclass[runningheads]{llncs}

\usepackage{graphicx}
\usepackage{amsmath}
\usepackage{amssymb}
\usepackage{mlmacros}
\usepackage{float}
\usepackage{verbatim}
\usepackage{booktabs}
\usepackage{afterpage}
\usepackage{nicefrac}
\usepackage{bm}
\usepackage{subfig}
\usepackage{caption}
\usepackage[absolute,overlay]{textpos}

\newcommand{\elbo}{\mathcal{L}}
\newcommand{\givn}{\nonscript\,\vert\nonscript\,}
\newcommand{\x}{\mathbf{x}}
\newcommand{\y}{\mathbf{y}}
\newcommand{\z}{\mathbf{z}}
\newcommand{\J}{\mathbf{J}}
\DeclareMathOperator{\MF}{MF}
\DeclareMathOperator{\I}{I}

\graphicspath{{figures/}}

\begin{document}

\title{Increasing the Generalisation Capacity of Conditional VAEs}

\author{Alexej Klushyn\thanks{Correspondence to: alexej.klushyn@argmax.ai} \and
	Nutan Chen \and
	Botond Cseke \and
	Justin Bayer \and
	Patrick van der Smagt}

\authorrunning{A. Klushyn et al.}

\institute{Machine Learning Research Lab, Volkswagen Group, Munich, Germany}

\maketitle

\begin{textblock*}{15cm}(5.2cm,25cm) 
	28th International Conference on Artificial Neural Networks, 2019
\end{textblock*}

\input{sections/abstract}
\input{sections/introduction}
\input{sections/methods}
\input{sections/related_work}
\input{sections/experiments}
\input{sections/conclusion}


\bibliographystyle{splncs04}
\bibliography{MMP_CVAE}

\end{document}

%% file: sections/abstract.tex
\begin{abstract}
We address the problem of one-to-many mappings in supervised learning, where a single instance has many different solutions of possibly equal cost.
The framework of conditional variational autoencoders describes a class of methods to tackle such structured-prediction tasks by means of latent variables.
We propose to incentivise informative latent representations for increasing the generalisation capacity of conditional variational autoencoders.
To this end, we modify the latent variable model by defining the likelihood as a function of the latent variable only and introduce an expressive multimodal prior to enable the model for capturing semantically meaningful features of the data.
To validate our approach, we train our model on the Cornell Robot Grasping dataset, and modified versions of MNIST and Fashion-MNIST obtaining results that show a significantly higher generalisation capability.

\keywords{Structured Prediction \and Latent Variable Models \and Conditional Variational Autoencoders \and Empirical Bayes.}
\end{abstract}

%% file: sections/introduction.tex
\section{Introduction}
\label{sec:introduction}
The problem of approximating conditional probability distributions $p(\y \givn \x)$ is a central point in the field of supervised learning.
Although, learning a complex many-to-one mapping is straightforward if a sufficient amount of data is available~\cite{krizhevsky2012imagenet,simonyan2014very}, most methods fail when it comes to structured-prediction problems, where a distribution with multiple modes (one-to-many mapping) has to be modelled~\cite{Tang2013}.

Conditional variational autoencoders~(CVAEs)~\cite{Sohn2015} are a class of latent variable models for approximating one-to-many functions.
They define a lower bound on the intractable marginal likelihood by introducing a variational posterior distribution.
The learned generative model and the corresponding (approximate) posterior distribution of the latent variables provide a decoder/encoder pair that captures semantically meaningful features of the data.
In this paper we address the issue of learning informative encodings/latent representations with the goal of increasing the generalisation capacity of CVAEs.

In contrast to variational autoencoders~(VAEs)~\cite{Kingma2013,Rezende2014}, the decoder of CVAEs is a function of the latent variable \emph{and} the condition $\x$. 
Thus, the model is not incentivised to learn an informative latent representation.
To tackle this problem, we propose to apply a VAE-like decoder that depends only on the latent variable.
This modification requires that the model is capable of learning a rich encoding.
We follow the line of argument in~\cite{chen2016variational}---where the expressiveness of the generative model is increased by introducing a flexible prior---and show that a \emph{multimodal} prior substantially improves optimisation.

Building on that, we propose to apply a learnable mixture distribution as prior.
We show that the classical mixture of Gaussians prior suffers from focusing on outliers during optimisation causing a badly trained generative model.
Instead of learning the means and variances of the respective mixture components directly, we address this issue by introducing a Gaussian mixture prior, inspired by~\cite{Tomczak2017}, that is parameterised through both the encoder and the decoder, and evaluated at learned pseudo latent variables.

%% file: sections/methods.tex
\section{Methods}
\label{sec:methods}

\subsection{Preliminaries: Conditional VAEs}
\label{sec:methods-preliminaries}
In structured prediction problems each condition~$\x$ can be related to several targets~$\y$ (one-to-many mapping), which results in a multimodal conditional distribution $p_\theta(\y \givn \x)$.
Conditional-latent-variable models (CLVM), defined by
\begin{equation}
\begin{aligned}
\label{eq:clvm}
p_\theta(\y \givn \x) = \int p_\theta(\y \givn \x, \z)\, p_\theta(\z \givn \x)\, \mathrm{d}\z ,
\end{aligned}
\end{equation}
are capable of modelling multimodality by means of latent variables~$\z$.
However, in most cases the integral in Eq.~(\ref{eq:clvm}) is intractable.
Amortised variational inference~\cite{Kingma2013,Rezende2014} allows to address this issue by approximating $p_\theta(\y \givn \x)$ through maximising the evidence lower bound~(ELBO):
\begin{equation}
\begin{aligned}
\label{eq:elbo-cvae}
\log p_\theta(\y \givn \x) \geq \mathbb{E}_{q_\phi(\z \givn \x, \y)} \left[ \log \frac{p_\theta(\y \givn \x, \z)\, p_\theta(\z \givn \x)}{q_\phi(\z \givn \x, \y)} \right] \eqqcolon \elbo_{\text{ELBO}}(\theta, \phi),
\end{aligned}
\end{equation}
where the parameters of the approximate posterior $q_\phi(\z \givn \x, \y)$, the likelihood $p_\theta(\y \givn \x, \z)$, and the prior $p_\theta(\z \givn \x)$ are defined as neural-network functions of the conditioning variables.
This model is known as conditional variational autoencoder~(CVAE)~\cite{Sohn2015}.
Consequently, we will refer to the neural networks representing $q_\phi(\z \givn \x, \y)$ and $p_\theta(\y \givn \x, \z)$ as encoder and decoder, respectively.

\subsection{Incentivising Informative Latent Representations}
\label{sec:methods-ilr}
In the CVAE, the likelihood is conditioned on $\z$ \emph{and} $\x$.
Therefore, the model is not incentivised to learn an informative latent representation.
Rather, latent variables can be viewed as an assistance for enabling multimodality in $p_\theta(\y \givn \x)$.
For being able to fully exploit the generalisation capacity of CVAEs, we argue that an informative latent representation is necessary.
Thus, $\z$ determines $\y$ completely, i.e. the mutual information ${\I(\x\,;\,\y \givn \z )=0}$.
Following this line of argument, we obtain ${\x \perp \y \givn \z}$, and thus ${p_\theta(\y \givn \x, \z)=p_\theta(\y \givn \z)}$, leading to the following CLVM:
\begin{equation}
\begin{aligned}
\label{eq:clvm2}
p_\theta(\y \givn \x) = \int p_\theta(\y \givn \z)\, p_\theta(\z \givn \x)\, \mathrm{d}\z.
\end{aligned}
\end{equation}
This modification enforces the model to learn a richer latent representation because all the information given by the training data has to be encoded.

However, the model must also be capable of learning such a complex latent representation.
In case of CVAEs, the prior $p_\theta(\z \givn \x)$ is usually defined as a Gaussian distribution, leading to limited flexibility of the model, and hence to a worse generalisation, as addressed in~\cite{chen2016variational} and shown in Sec.~\ref{sec:experiments-gen} and \ref{sec:experiments-grasping}.
We build on the line of argumentation in~\cite{chen2016variational}, where the above limitation is tackled by introducing an expressive prior.
The KL-divergence $\mathop{\mathbb{KL}}\big(q_\phi(\z \givn \x, \y)\|~p_\theta(\z \givn \x)\big)$ in Eq.~\ref{eq:elbo-cvae} can be viewed as a regulariser to avoid over-fitting.
Therefore, a flexible prior allows for learning a more complex latent representation and leads automatically to a more expressive generative model $p_\theta(\y \givn \z)\, p_\theta(\z \givn \x)$.

\subsection{Modelling Low-Density Regions}
\label{sec:methods-zero}
In the previous section, we discussed the need of expressive priors in our setting.
Next, we will specify an important property the prior has to posses.
In most models within the VAE/CVAE framework, the prior is defined as a \emph{unimodal} distribution.
This leads to a significant shortcoming illustrated by the following structured-prediction task:
generating grasping poses~(targets) for a certain object~(condition).
Imagine a generated grasping pose is located in the middle of a plate instead of on the edge.
Hence, generating targets between modes of $p_\theta(\y \givn \x)$ might be an exclusion criterion.

To understand the cause, let us assume a dataset consisting of only a single condition with different targets.
Thus, $q_\phi(\z \givn \x, \y)=q_\phi(\z \givn \y)$, $p_\theta(\y \givn \x)=p_\theta(\y)$, and $p_\theta(\z \givn \x)=p_\theta(\z)$~(note that this is equivalent to a vanilla VAE).
We want to represent $p_\theta(\y)$ by transforming $p_\theta(\z)$ through a bijective function $g(\cdot)$, i.e.\ $\y = g(\z)$.
By applying the change of variables, we derive:
\begin{equation*}
\begin{aligned}
p_\theta(g(\z)) = 
\frac{1}{\sqrt{\det(\J^T\,\J)}}\, p_\theta(\z), ~~~\text{with}~~~ \J=\frac{\partial g(\z)}{\partial\z}.
\end{aligned}
\end{equation*}
In this context, we define the magnification factor $\MF\coloneqq\sqrt{\det(\J^T\,\J)}$~\cite{bishop1997magnification}. 
Setting $p_\theta(g(\z))=0$ requires either $p_\theta(\z)=0$ or $\MF\rightarrow \infty$.
Thus, zero-density regions can only be represented at $\y$ if either the original density is zero or the $\MF$ becomes infinitely large~(see Sec.~\ref{sec:experiments-vis} for visualisation).
For example, when using a Gaussian distribution as prior, near-zero density regions occur only at its tails.
If $g(\cdot)$ is the likelihood neural network and we assume it to be continuous, zero-density regions can only be obtained in tails. 
For zero densities elsewhere, infinitely large $\MF$-values are required.  
Thus, the derivative of $g(\cdot)$ becomes infinitely large: $\J \rightarrow \infty$, leading to a badly-conditioned optimisation problem.
The above line of argument applies equally to datasets with multiple conditions.

\subsection{Expressive Priors for Conditional VAEs}
\label{sec:methods-cdvp}
A natural approach to address the difficulties introduced in Sec.~\ref{sec:methods-ilr} and \ref{sec:methods-zero} is a flexible multimodal prior.
This could be realised by a conditional mixture of Gaussians~(CMoG) prior ${p_\theta(\z \givn \x) = \frac{1}{K} \sum_{k=1}^{K} \mathcal{N}\big(\mu_k (\x), \text{diag}(\sigma_k^2(\x))\big)}$,
where $K$ is the number of mixture components.
As in case of the vanilla CVAE, the parameters of the prior ${\big\{\mu_{k}(\x), \text{diag}(\sigma_k^2(\x))\big\}_{k=1}^K \in \mathbb{R}^{N_\z}}$ are represented by a neural network.
Unfortunately, this approach performs badly, especially in high dimensional latent spaces~(see Sec.~\ref{sec:experiments-gen}).

We suspect this mainly due to the following reason: 
the prior is optimised through minimising $\mathop{\mathbb{KL}}\big(q_\phi(\z \givn \x, \y)\|~p_\theta(\z \givn \x)\big)$~(see Eq.~\ref{eq:elbo-cvae}).
The optimal Bayes prior is the aggregated posterior ${p^{*}(\z \givn \x)=\mathop{\mathbb{E}_{\y\sim\hat{p}(\x, \y)}} q_\phi(\z \givn \x, \y)}$---representing the manifold of the encoded data.
Since the parameters of each mixture component of the CMoG prior are learned independently, it is not possible to avoid that mixture components leave the manifold of the encoded data by focusing on outliers~(see Sec.~\ref{sec:experiments-gen} for experimental support).
This leads to a badly trained generative model.
Thus, the problem is that the prior is not incentivised to stay on the manifold of the encoded data.

Instead of learning the mean and variance of each mixture component of the prior directly, we tackle the above issue by introducing a parameterisation through both the encoder and the decoder.
This approach is inspired by the VampPrior~\cite{Tomczak2017} (VAE framework), which is parameterised through the encoder.
When extending it to the CVAE framework, we obtain the conditional VampPrior $p(\z \givn \x) = \frac{1}{K} \sum_{k=1}^{K} q_\phi\big(\z \givn \x, \tilde{\mathbf{y}}_{k} \big)$, which is evaluated at learned pseudo targets ${\big\{\tilde{\mathbf{y}}_{k}\big\}_{k=1}^K \in \mathbb{R}^{N_\y}}$.
However, pseudo latent variables $\tilde{\mathbf{z}}$ would require less parameters and thus are less complex to optimise for representing the manifold of the encoded data.
Evaluating the conditional VampPrior at decoded $\tilde{\mathbf{z}}$ would make use of this advantage~(see Sec.~\ref{sec:experiments-gen} for experimental support).
Below, we introduce the conditional decoder-based Vamp~(CDV)~prior:
\begin{equation}
\begin{aligned}
\label{eq:CDV}
p_{\pi}(\z \givn \x) = \frac{1}{K} \sum_{k=1}^{K} q_\phi\big(\z \givn \x, \mu_\theta(\tilde{\mathbf{z}}_{k}(\x)) \big) ,
\end{aligned}
\end{equation}
where $\mu_\theta(\cdot)$ is the mean of the likelihood and ${\big\{\tilde{\mathbf{z}}_{k}(\x)\big\}_{k=1}^K \in \mathbb{R}^{N_\z}}$ are defined as functions of the condition and approximated by a single neural network $f_\psi(\x)$, which is trained through backpropagation.
Thus, the parameters of the prior are $\pi = \{\psi, \theta, \phi\}$.
As an additional feature, this approach requires less parameters than the CMoG prior, since only the pseudo latent variables~(${\in \mathbb{R}^{N_\z}}$)) have to be learned instead of the means and variances (each ${\in \mathbb{R}^{N_\z}}$) of the CMoG prior.

The CLVM in Eq.~\ref{eq:clvm2} was introduced to incentivise a more informative latent representation for achieving a higher generalisation capacity. 
This step demands a flexible multimodal prior that allows the model for capturing semantically meaningful features of the data.
The CDV prior meets these requirements and, in contrast to a classical Gaussian mixture prior, it facilitates a well trained generative model. 

%% file: sections/related_work.tex
\section{Related Work}
\label{sec:related_work}
Learning informative latent representations in VAEs is an ongoing field of research~\cite{higgins2017beta,sonderby2016ladder,alemi2017fixing}.
The connection between informative latent representations and a flexible prior was pointed out in~\cite{chen2016variational} and motivated through Bits-Back Coding.
Several additional works improved VAEs by learning more complex priors~\cite{nalisnick2016stick,Tomczak2017}.
The reason for increasing the expressiveness of the prior is a lower KL-divergence---and thus a better trained decoder, leading to more qualitative samples of the generative model.
Based on that, it can be derived that the optimal Bayes prior is the aggregated posterior~\cite{Tomczak2017}.
The VampPrior~\cite{Tomczak2017} approximates the aggregated posterior by a uniform mixture of approximate posteriors, evaluated at learned pseudo inputs in the observable space.

In contrast to the (conditional) VampPrior, the CDV prior is parameterised through both the encoder and the decoder, and evaluated at learned pseudo latent variables.
Since the latent space has in general a lower dimension than the observable space, pseudo latent variables need less parameters and are easier to optimise for approximating the aggregated posterior.

Several applications based on the concept of CVAEs were published:
they can be used for filling pixels given a partial image~\cite{Sohn2015}, for image inpainting conditioned on visual attributes~(e.g., colour and gender)~\cite{yan2016attribute2image}, or for predicting events by conditioning the distribution of possible movements on a scene~\cite{walker2016uncertain}. 
As in~\cite{Sohn2015}, we use CVAEs to complete images---with the aim of obtaining a widest possible variety of generations, thus a classical one-to-many mapping.
However, with an additional difficulty:
it is learned from a dataset of one-to-one mappings to validate the generalisation capacity of the models.

Another important field where CVAEs are applied is robot grasping:
earlier work has focused on detecting robust grasping poses~\cite{lenz2015,pinto2016supersizing}, while recent work is often based on structured prediction with the idea of learning multimodal conditional probability distributions for generating grasping poses~\cite{veres2017modeling}.
In~\cite{lenz2015,pinto2016supersizing}, classifiers are applied to detect whether a grasping pose is robust. 
A problem here is that suitable grasping poses need to be proposed by hand.
In our approach, CVAEs are used to generate grasping poses for unknown objects.
Afterwards, similar to~\cite{lenz2015}, a discriminator is applied to validate them.

%% file: sections/experiments.tex
\section{Experiments}
\label{sec:experiments}
We conduct five experiments to compare the introduced models:
first, we visualise on a simplified task the difficulty of unimodal priors.
Building on that, we demonstrate on a synthetic toy dataset that CMoG- and CDV-CVAEs are capable of modelling near-zero-density regions.
Second, we show on a modified version of MNIST and Fashion-MNIST that the variety of generated samples is significantly larger when combining the CVAE with the CMoG or CDV prior.
Finally, we compare the CVAE with the CDV-CVAE on real world data, the Cornell Robot Grasping dataset.

To train our models we applied a linear annealing scheme~\cite{bowman2015generating} for the first epoch.
This is especially important for the CDV-CVAE because it is sensitive to over-regularisation by the KL-term in the initial optimisation phase.

\subsection{Modelling Low-Density Regions}
\label{sec:experiments-zero}

\subsubsection{Visualisation of the Problem}
\label{sec:experiments-vis}
\begin{figure*}[t]
	\centering
	\subfloat[Latent representation of four Gaussians\label{fig:5-1a}]{\includegraphics[width=0.55\textwidth]{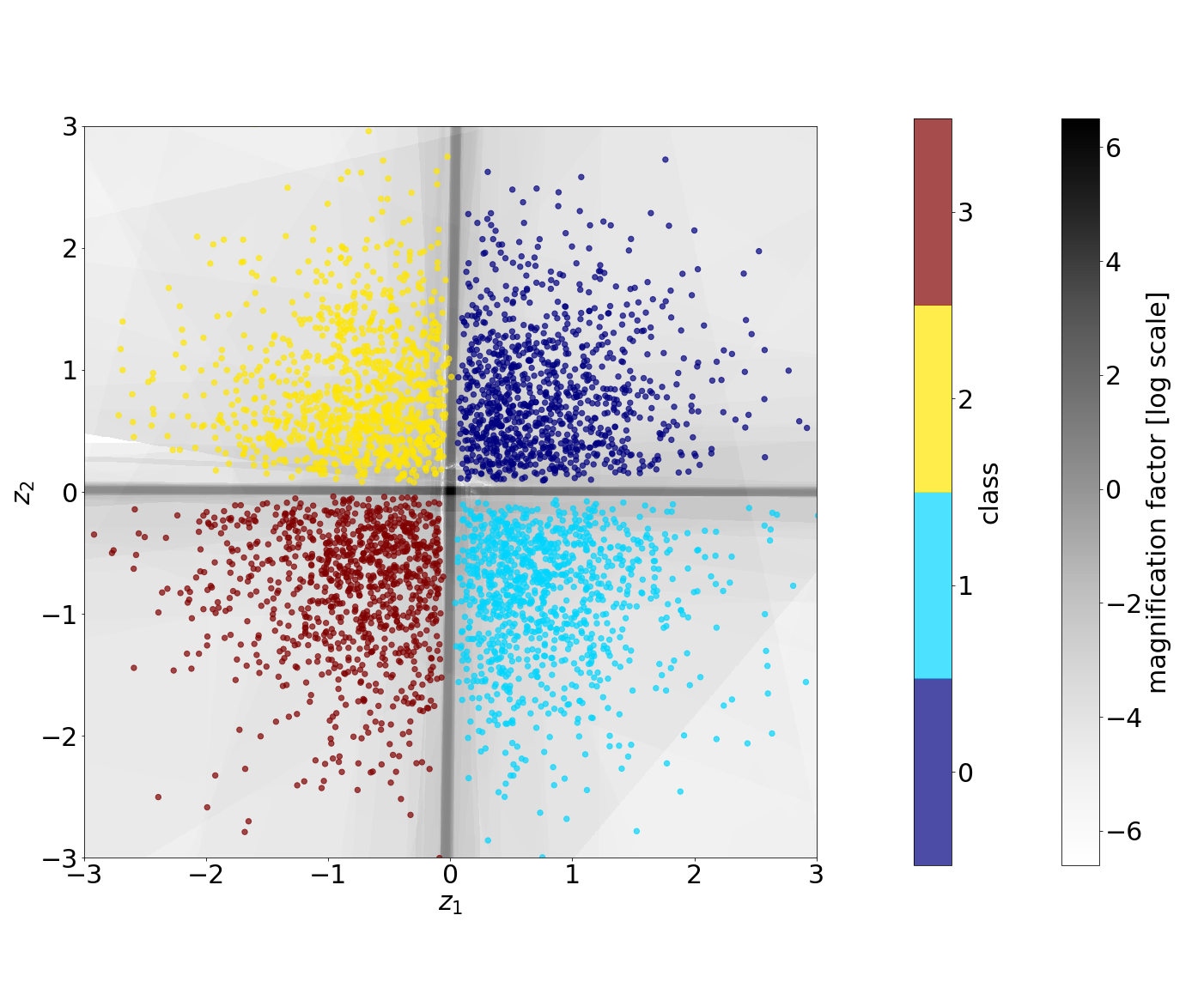}}\hfill
	\subfloat[Generated samples\label{fig:5-1b}] {\includegraphics[width=0.45\textwidth]{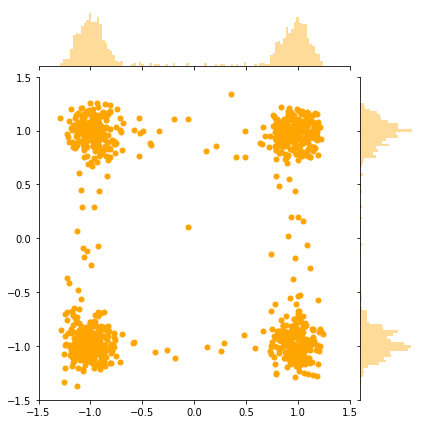}}\hfill
	\caption{%
		Effect of unimodal priors on the performance of VAEs/CVAEs.
		for illustration, we use a dataset of four Gaussian distributions arranged in a square.	
		Latent representation (a):
		the colours encode the four different Gaussians.
		The greyscale indicates the gradients of the decoder, which are required to map from a unimodal to a multimodal distribution. 
		1,000 generated samples (b):
		we also obtain samples between modes, since the decoder is a continuous function approximated by a neural network.
		(see Sec.~\ref{sec:experiments-vis})
	}%
	\label{fig:5-1}
\end{figure*}
To reduce complexity, we trained a vanilla VAE with a Gaussian prior on a simple toy dataset consisting of four Gaussian distributions.
This toy dataset can be interpreted as a simplified structured-prediction task with only one condition and four targets.

Fig.~\ref{fig:5-1a} shows the two-dimensional latent space, which depicts the aggregated posterior of the model. 
Each of the four Gaussians is encoded by a different colour.
To map from a unimodal to a multimodal distribution, the decoder has to model large gradients, as discussed in Sec.~\ref{sec:methods-zero}.
The magnification factor is visualised by the greyscale in Fig.~\ref{fig:5-1a}, which represents the Jacobian of the decoder.
The support of the aggregated posterior is noticeably smaller than the support of the prior.
Since the decoder is a continuous function, a gap at the boundaries of different classes in the latent space (as shown in Fig.~\ref{fig:5-1a}) represents the distance between the modes in the observable space. 
The size of the gap depends on the gradients that our model is able to achieve:
the higher the gradient, the smaller the gap in the latent space.

When sampling from the generative model, we first sample from the prior.
If the sample comes from a region which is not supported by the aggregated posterior, the decoded sample will end up between two modes, as demonstrated in Fig.~\ref{fig:5-1b}.

\subsubsection{Synthetic Toy Dataset}
\label{sec:experiments-toy}
\begin{figure*}[t]
	\centering
	\subfloat[Training data \label{fig:5-2a}] {\includegraphics[width=0.25\textwidth]{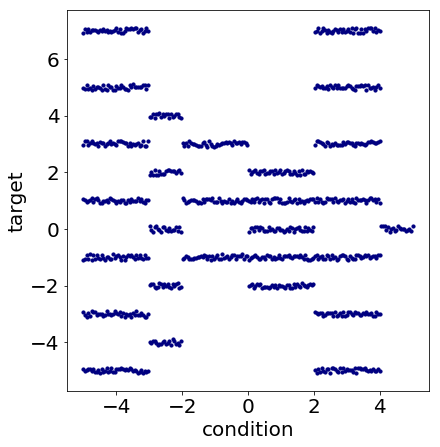}}\hfill
	\subfloat[CVAE \label{fig:5-2b}] {\includegraphics[width=0.25\textwidth]{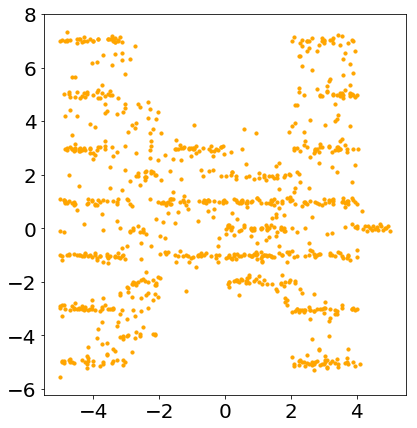}}\hfill
	\subfloat[CMoG-CVAE \label{fig:5-2c}] {\includegraphics[width=0.25\textwidth]{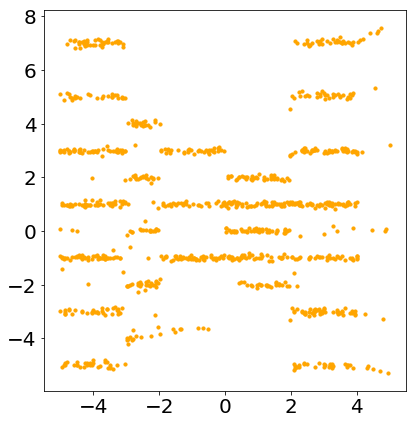}}\hfill
	\subfloat[CDV-CVAE \label{fig:5-2d}] {\includegraphics[width=0.25\textwidth]{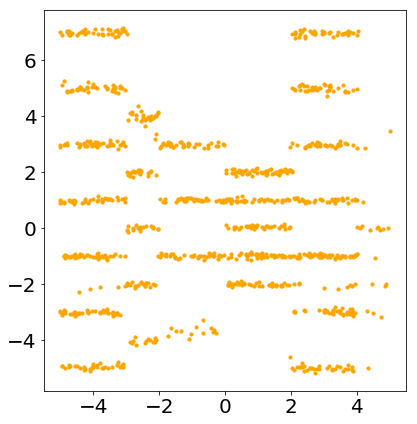}}\hfill
	\caption{%
		Synthetic toy dataset~(a) of one-dimensional one-to-many mappings.
		The horizontal axis represents the conditions, the vertical axis the targets.
		Generated samples~(b--d):
		a near-zero-density between different modes is only achieved through multimodal priors, as shown in~(c) and (d).
		(see Sec.~\ref{sec:experiments-toy})
	}%
	\label{fig:5-2}
\end{figure*}
\begin{figure*}[t]
	\centering	
	\subfloat[$\x = 0$ \label{fig:5-3a}] {\includegraphics[width=0.25\textwidth]{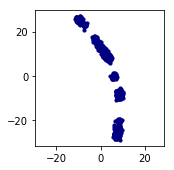}}\hfill
	\subfloat[$\x = 1$ \label{fig:5-3b}] {\includegraphics[width=0.25\textwidth]{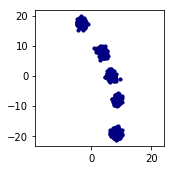}}\hfill
	\subfloat[$\x = 2$ \label{fig:5-3c}] {\includegraphics[width=0.25\textwidth]{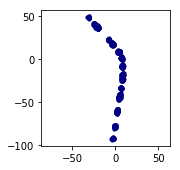}}\hfill
	\subfloat[$\x = 3$ \label{fig:5-3d}] {\includegraphics[width=0.25\textwidth]{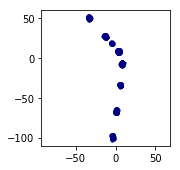}}\hfill
	\caption{%
		Samples from the CDV prior depending on the condition $\x$ (trained on the synthetic toy dataset Fig.~\ref{fig:5-2a}).
		The number of modes of the prior and of the likelihood distribution are similar~(see Fig.~\ref{fig:5-2d}).
		If the number of targets changes, the prior modes merge, as shown in (a) and (c).
		(see Sec.~\ref{sec:experiments-toy})	
	}%
	\label{fig:5-3}
\end{figure*}
In this experiment we reused a synthetic toy dataset~\cite{Tang2013} for validating models for structured-prediction tasks.
It consists of one-dimensional one-to-many mappings~(see Fig.~\ref{fig:5-2a}):
the horizontal-axis represents the conditions and the vertical-axis the targets.
Even though the dataset is simple, the abrupt changes of the number and location of the targets are quite challenging to model. 

For all three models, we used latent spaces with two dimensions.
CMoG-CVAE~(${\elbo_{\text{ELBO}}=-0.586}$) and CDV-CVAE~(${\elbo_{\text{ELBO}}=-0.518}$) outperformed the original CVAE~(${\elbo_{\text{ELBO}}=-1.12}$) as shown in Fig.~\ref{fig:5-2}.
Multimodal priors facilitate the modelling of near-zero-density regions between different modes~(Fig.~\ref{fig:5-2c}, \ref{fig:5-2d}), as discussed in Sec.~\ref{sec:methods-zero}.
Fig.~\ref{fig:5-3} shows how the CDV prior distribution changes with the condition $\x$.

\subsection{Verifying the Generalisation Capacity}
\label{sec:experiments-gen}
\begin{figure*}[t]
	\centering	
	{\includegraphics[width=0.325\textwidth]{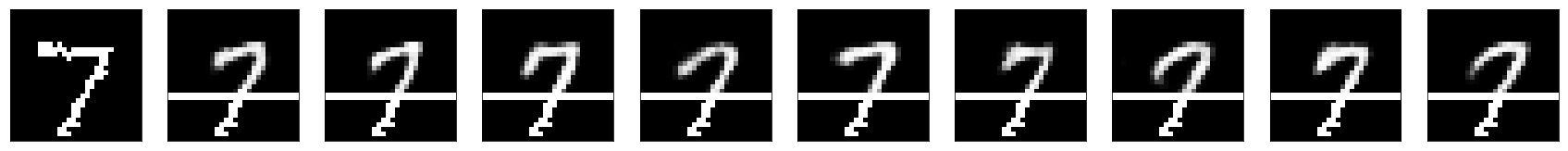}}\hfill%
	{\includegraphics[width=0.325\textwidth]{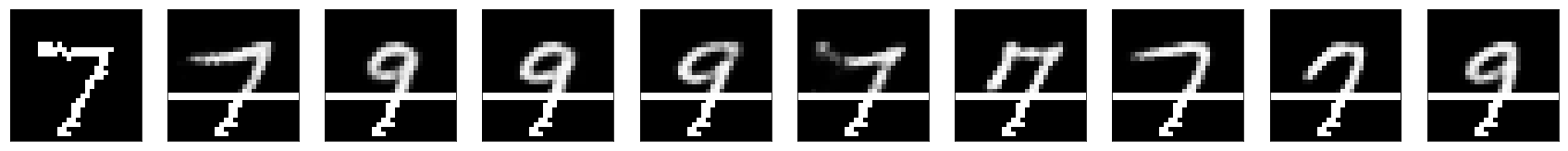}}\hfill%
	{\includegraphics[width=0.325\textwidth]{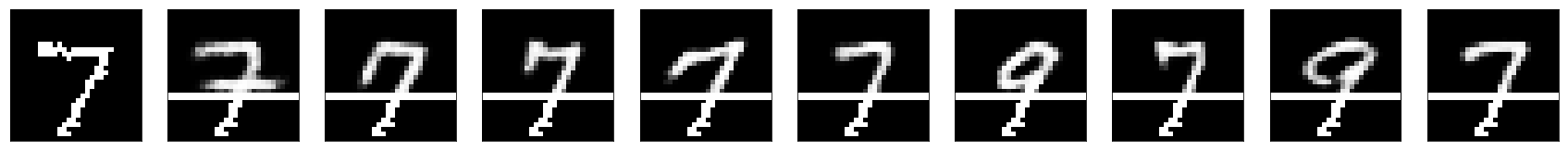}}\hfill%
	{\includegraphics[width=0.325\textwidth]{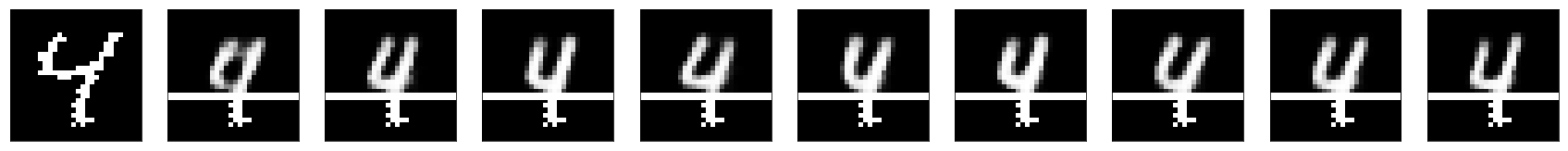}}\hfill%
	{\includegraphics[width=0.325\textwidth]{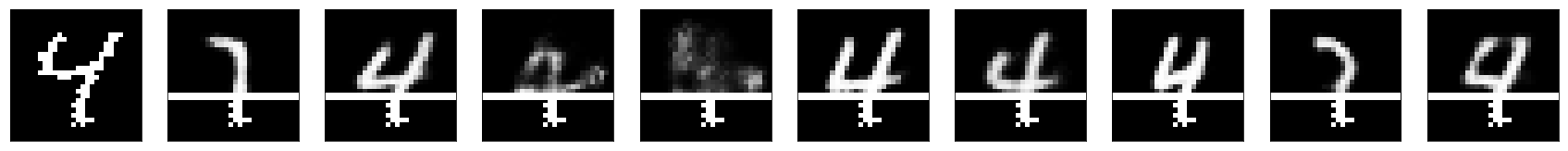}}\hfill%
	{\includegraphics[width=0.325\textwidth]{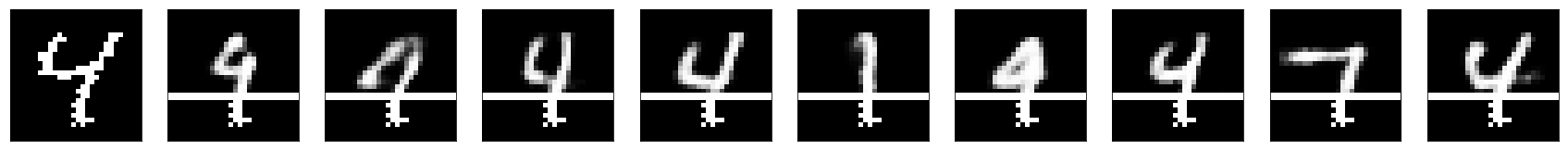}}\hfill%
	{\includegraphics[width=0.325\textwidth]{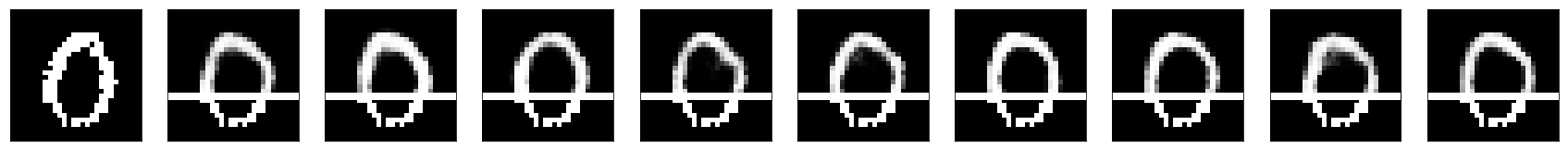}}\hfill%
	{\includegraphics[width=0.325\textwidth]{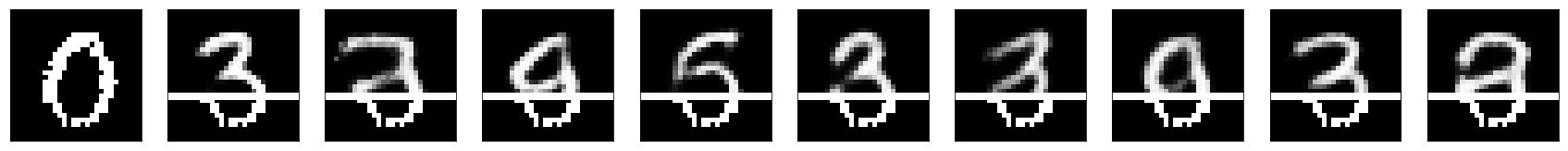}}\hfill%
	{\includegraphics[width=0.325\textwidth]{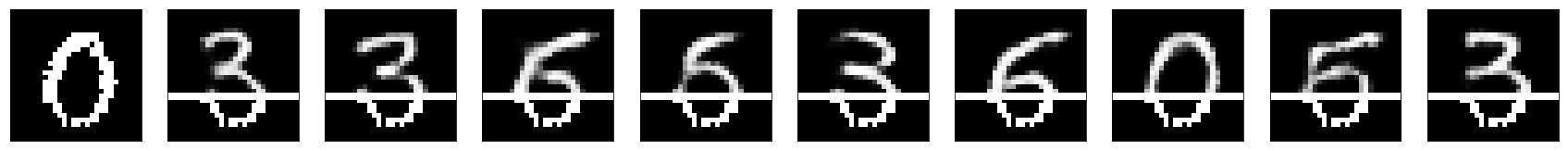}}\hfill%
	{\includegraphics[width=0.325\textwidth]{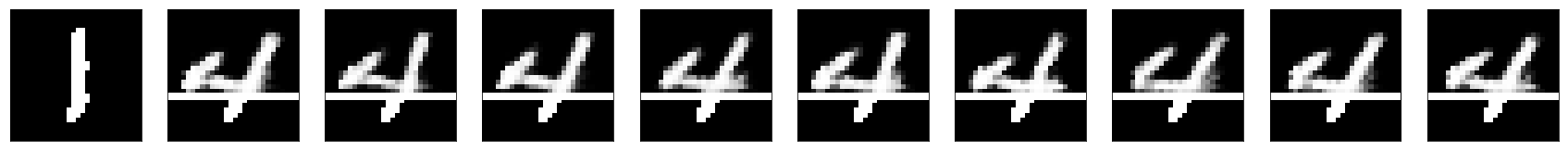}}\hfill%
	{\includegraphics[width=0.325\textwidth]{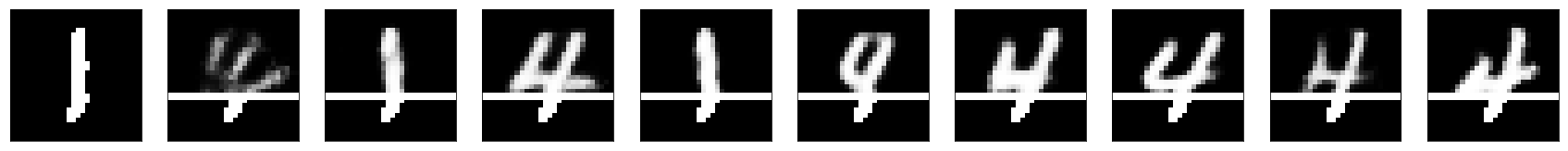}}\hfill%
	{\includegraphics[width=0.325\textwidth]{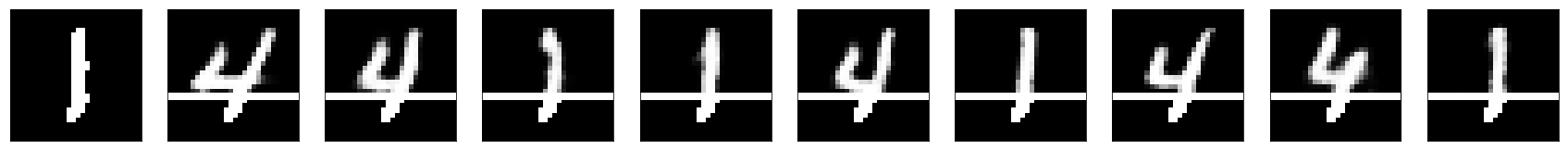}}\hfill%
	{\includegraphics[width=0.325\textwidth]{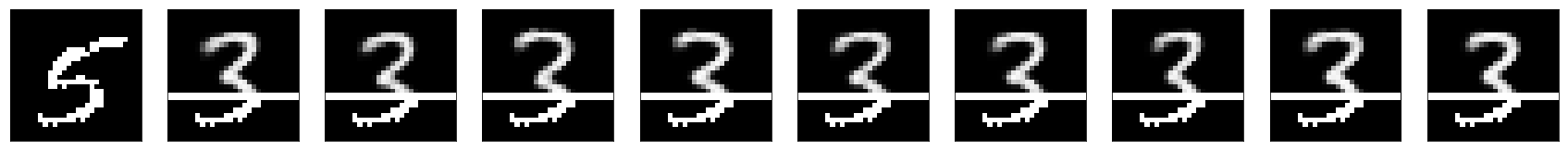}}\hfill%
	{\includegraphics[width=0.325\textwidth]{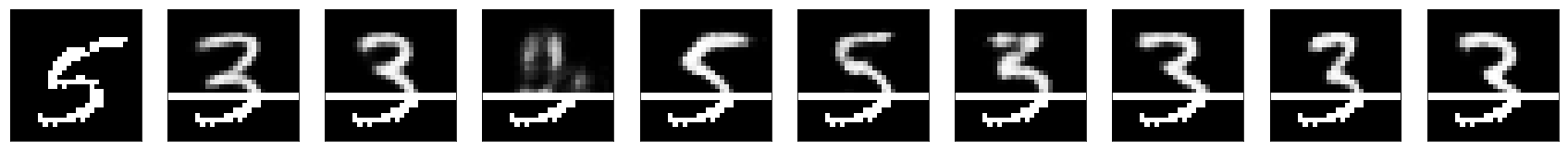}}\hfill%
	{\includegraphics[width=0.325\textwidth]{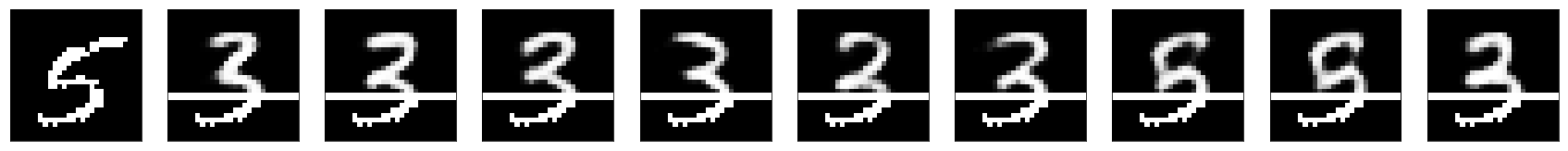}}\hfill%
	{\includegraphics[width=0.325\textwidth]{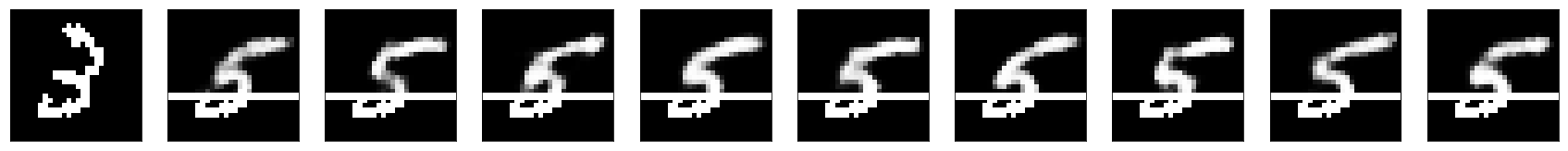}}\hfill%
	{\includegraphics[width=0.325\textwidth]{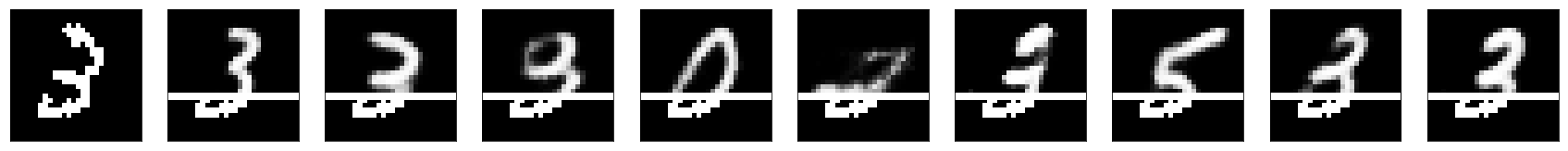}}\hfill%
	{\includegraphics[width=0.325\textwidth]{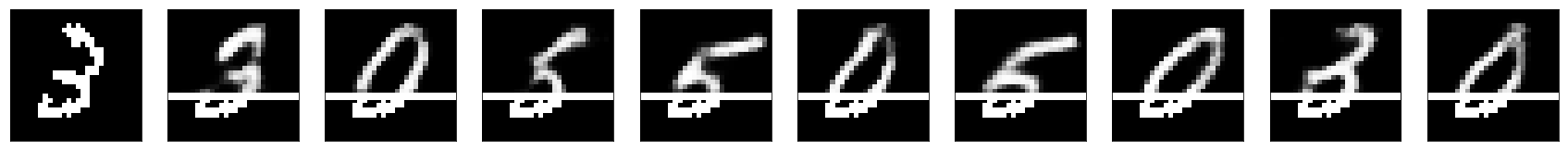}}\hfill%
	{\includegraphics[width=0.325\textwidth]{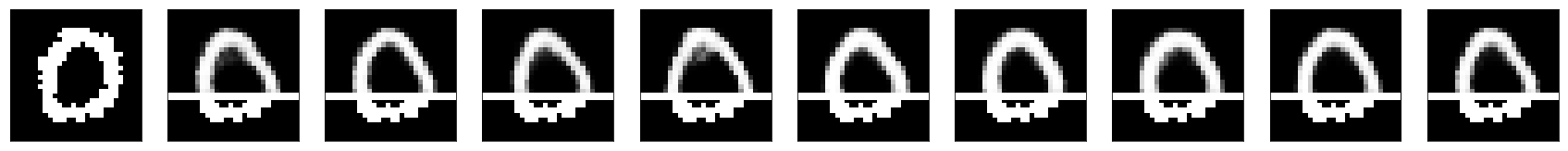}}\hfill%
	{\includegraphics[width=0.325\textwidth]{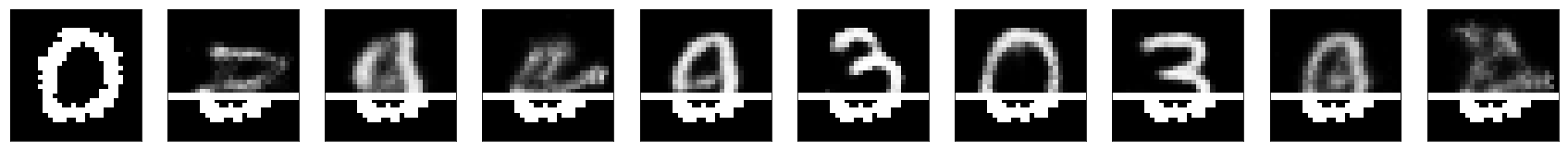}}\hfill%
	{\includegraphics[width=0.325\textwidth]{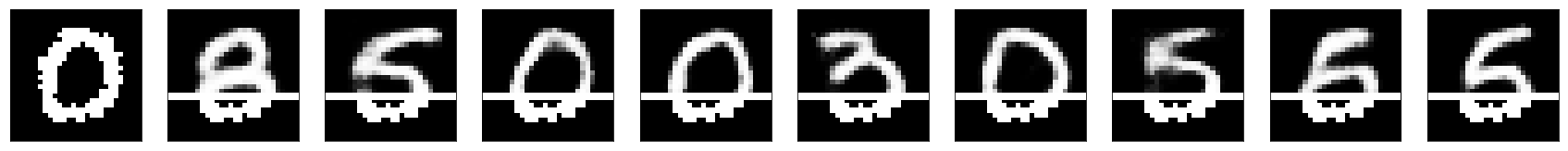}}\hfill%
	{\includegraphics[width=0.325\textwidth]{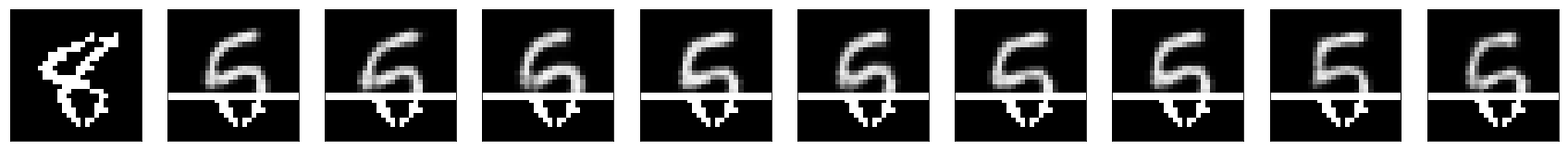}}\hfill%
	{\includegraphics[width=0.325\textwidth]{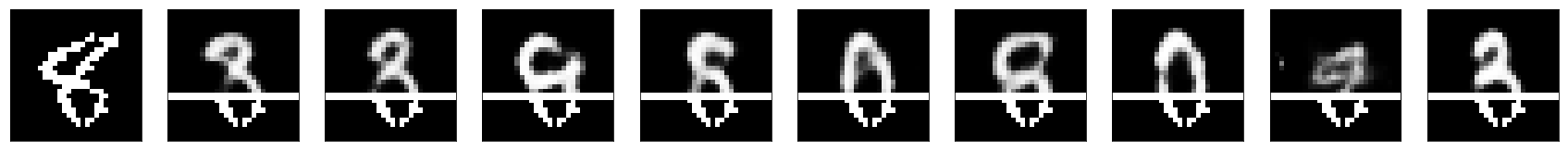}}\hfill%
	{\includegraphics[width=0.325\textwidth]{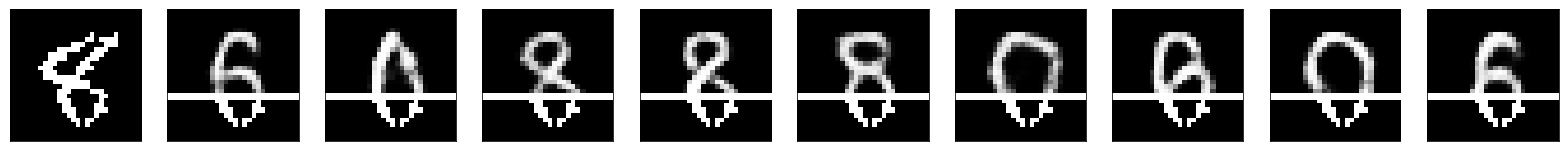}}\hfill%
	{\includegraphics[width=0.325\textwidth]{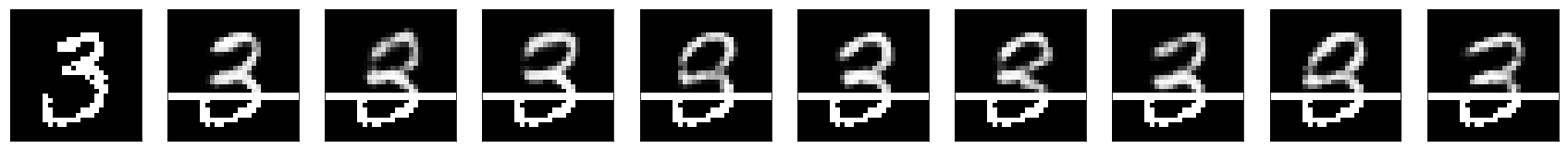}}\hfill%
	{\includegraphics[width=0.325\textwidth]{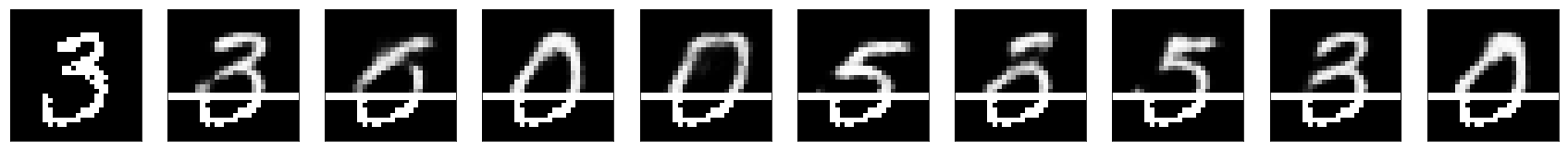}}\hfill%
	{\includegraphics[width=0.325\textwidth]{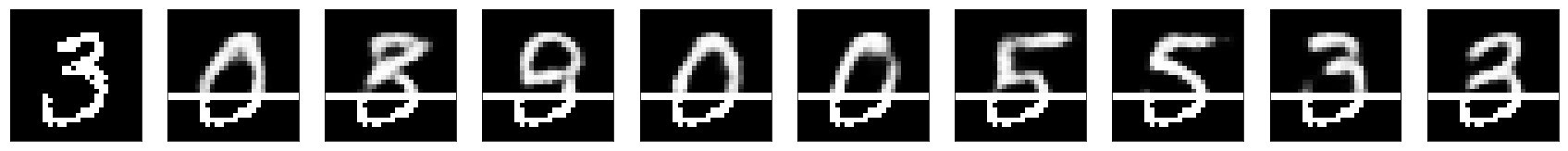}}\hfill%
	{\includegraphics[width=0.325\textwidth]{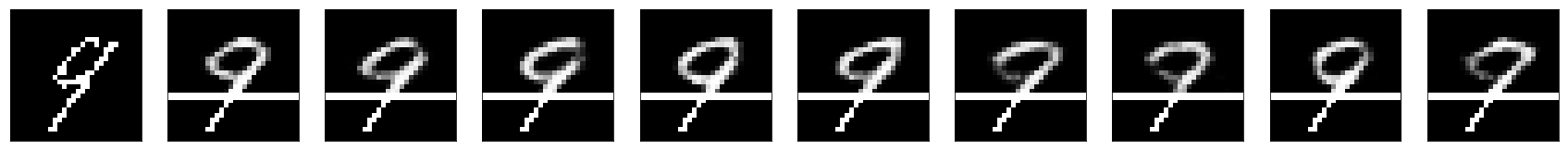}}\hfill%
	{\includegraphics[width=0.325\textwidth]{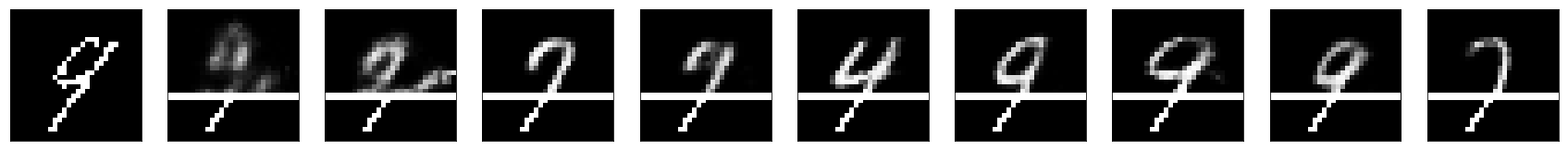}}\hfill%
	{\includegraphics[width=0.325\textwidth]{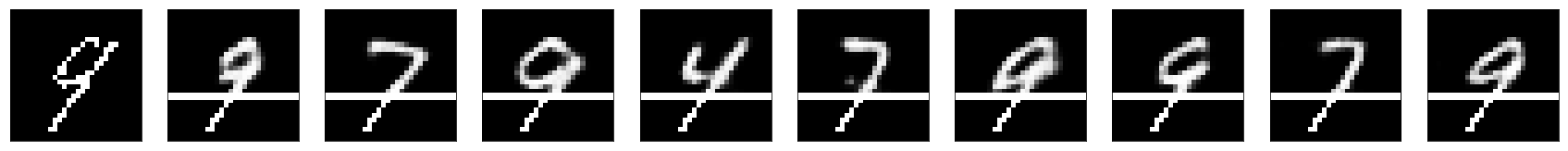}}\hfill%
	\vskip +5pt
	{\includegraphics[width=0.325\textwidth]{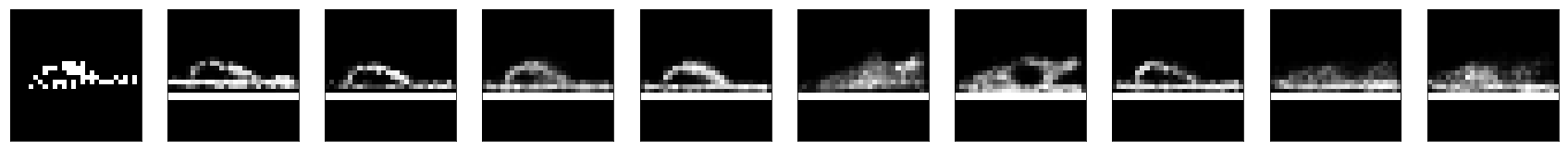}}\hfill%
	{\includegraphics[width=0.325\textwidth]{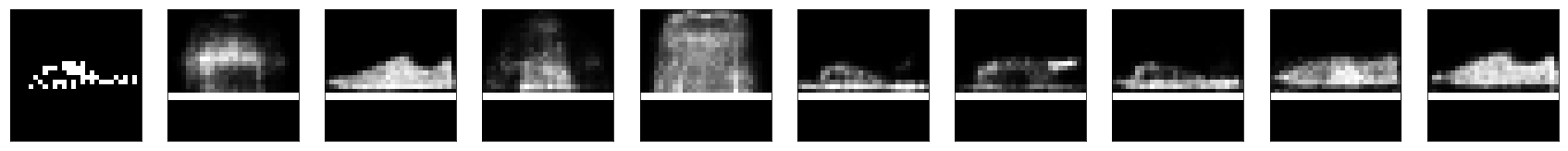}}\hfill%
	{\includegraphics[width=0.325\textwidth]{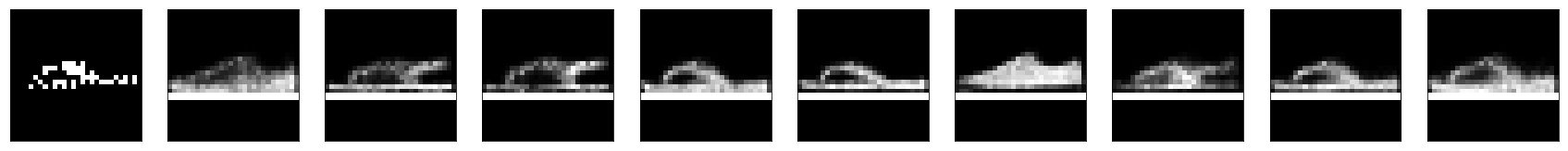}}\hfill%
	{\includegraphics[width=0.325\textwidth]{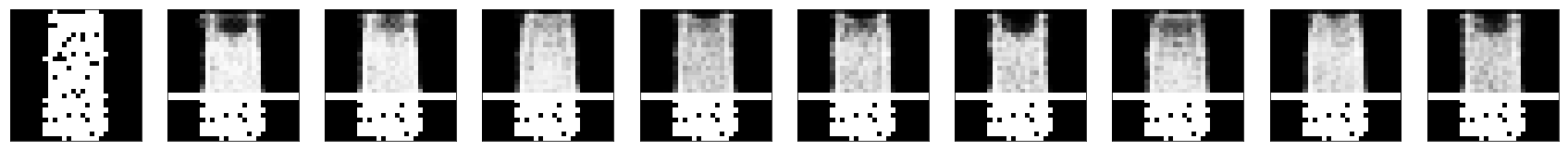}}\hfill%
	{\includegraphics[width=0.325\textwidth]{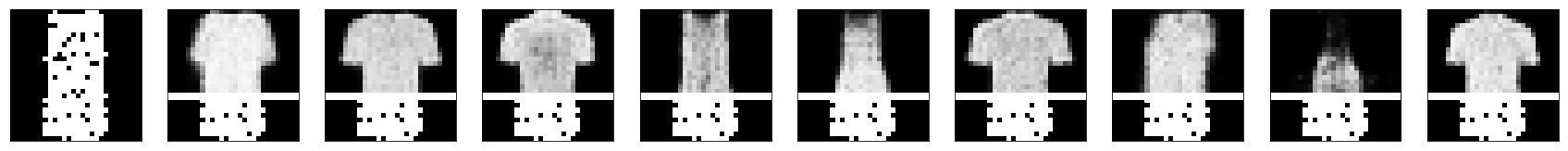}}\hfill%
	{\includegraphics[width=0.325\textwidth]{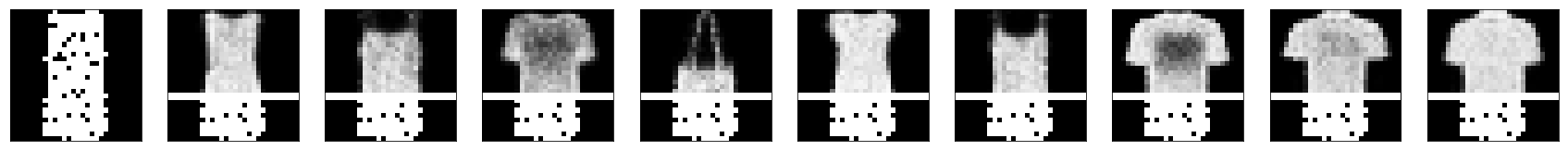}}\hfill%
	{\includegraphics[width=0.325\textwidth]{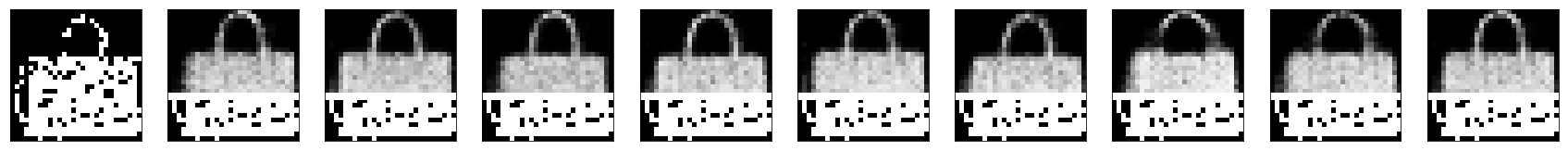}}\hfill%
	{\includegraphics[width=0.325\textwidth]{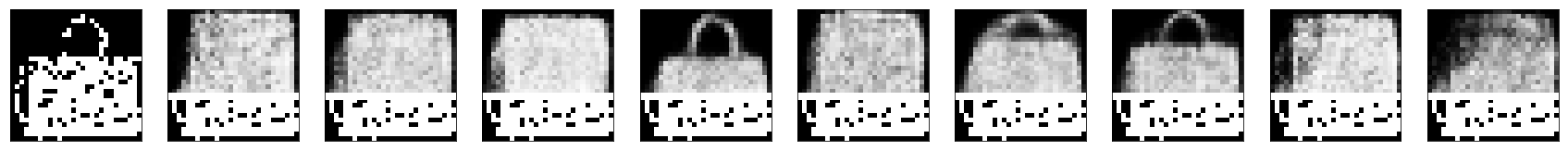}}\hfill%
	{\includegraphics[width=0.325\textwidth]{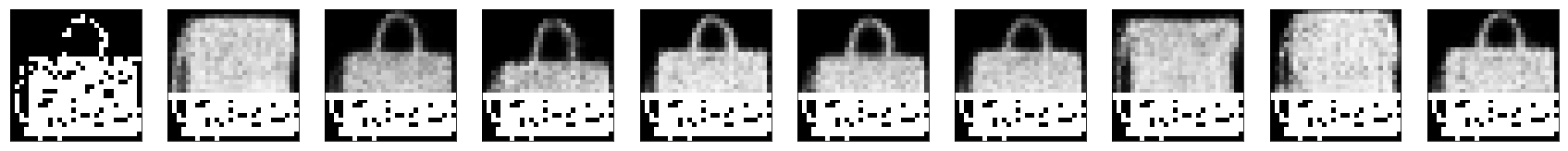}}\hfill%
	{\includegraphics[width=0.325\textwidth]{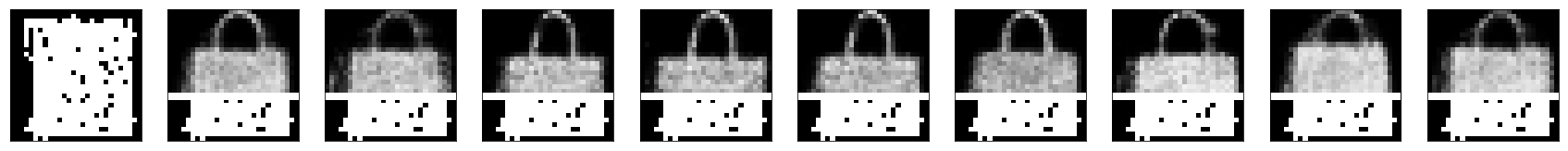}}\hfill%
	{\includegraphics[width=0.325\textwidth]{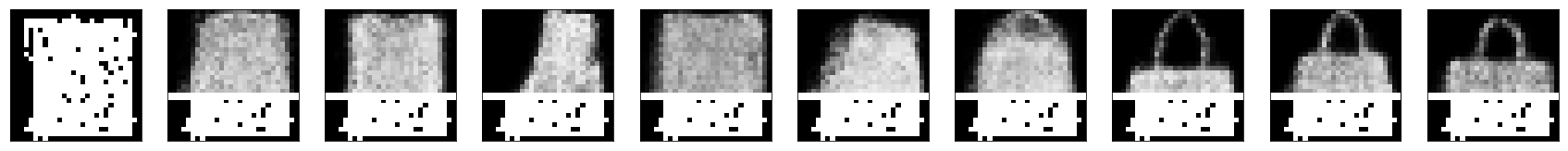}}\hfill%
	{\includegraphics[width=0.325\textwidth]{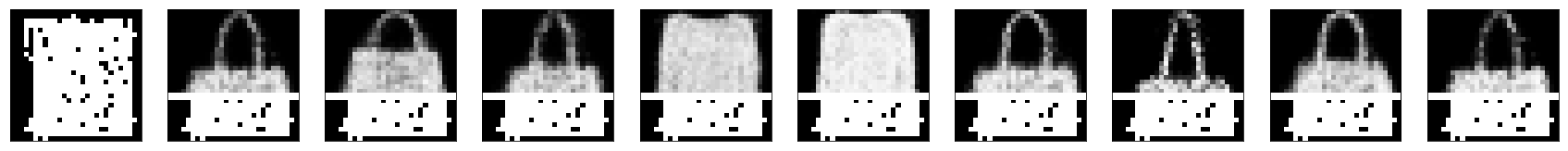}}\hfill%
	{\includegraphics[width=0.325\textwidth]{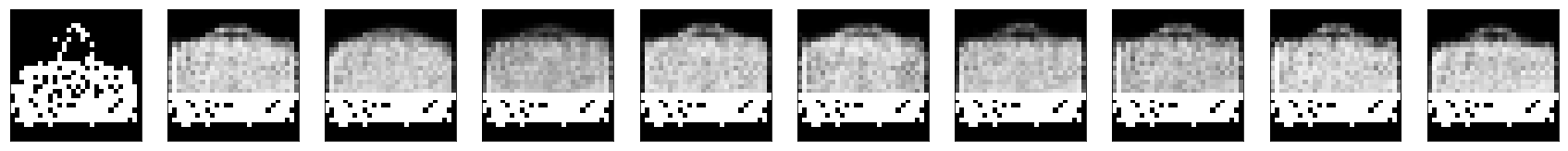}}\hfill%
	{\includegraphics[width=0.325\textwidth]{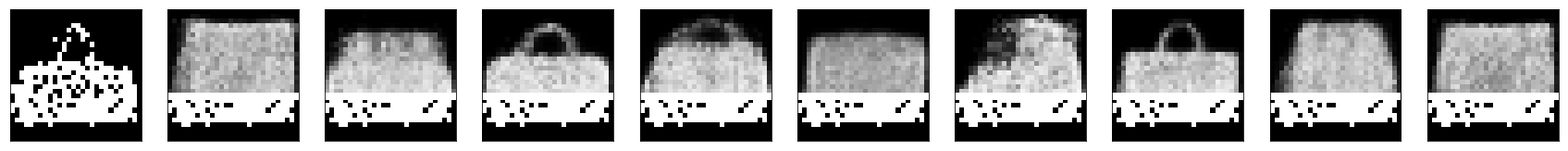}}\hfill%
	{\includegraphics[width=0.325\textwidth]{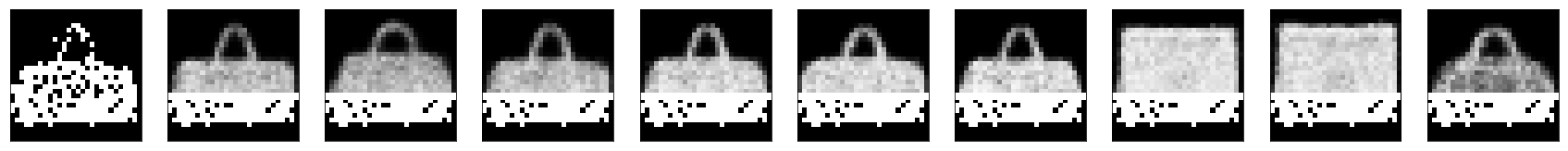}}\hfill%
	{\includegraphics[width=0.325\textwidth]{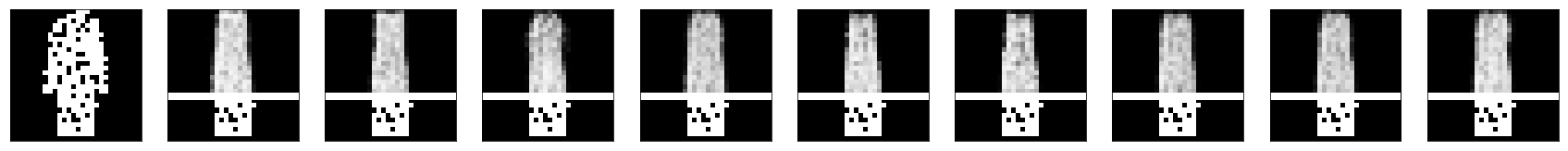}}\hfill%
	{\includegraphics[width=0.325\textwidth]{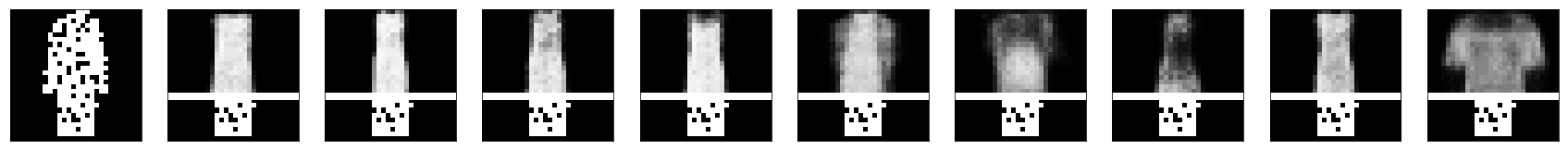}}\hfill%
	{\includegraphics[width=0.325\textwidth]{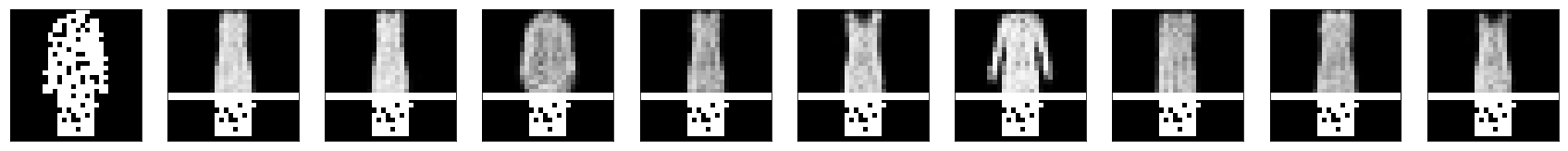}}\hfill%
	{\includegraphics[width=0.325\textwidth]{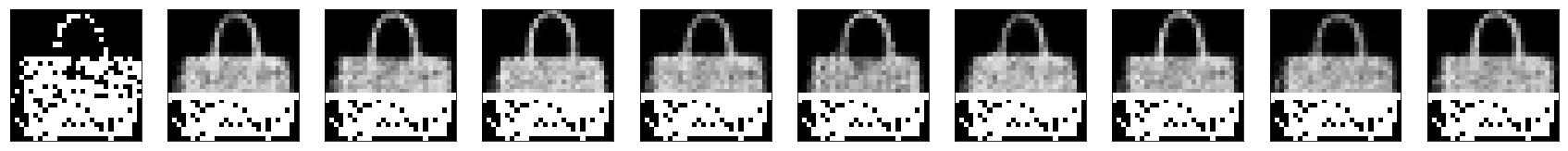}}\hfill%
	{\includegraphics[width=0.325\textwidth]{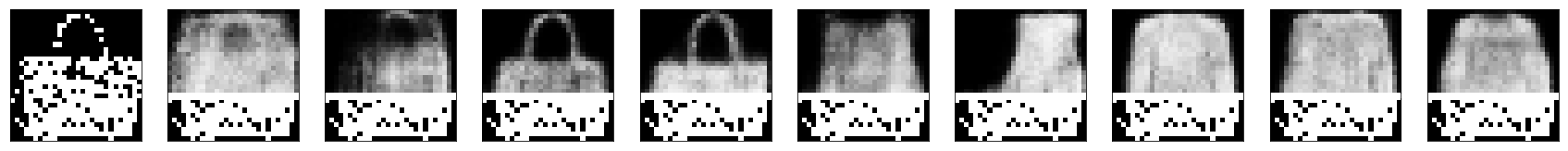}}\hfill%
	{\includegraphics[width=0.325\textwidth]{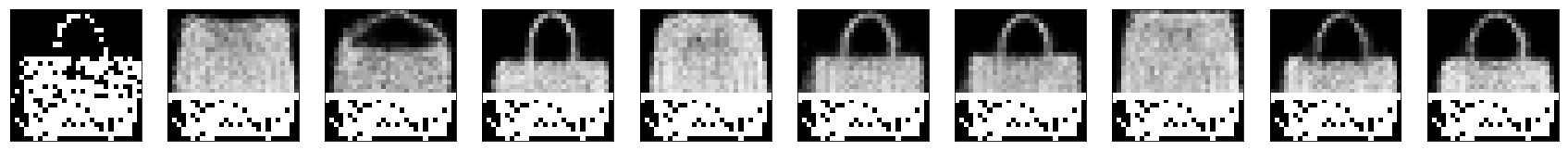}}\hfill%
	{\includegraphics[width=0.325\textwidth]{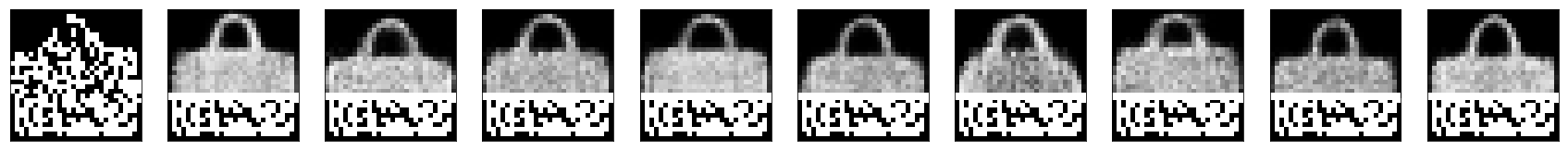}}\hfill%
	{\includegraphics[width=0.325\textwidth]{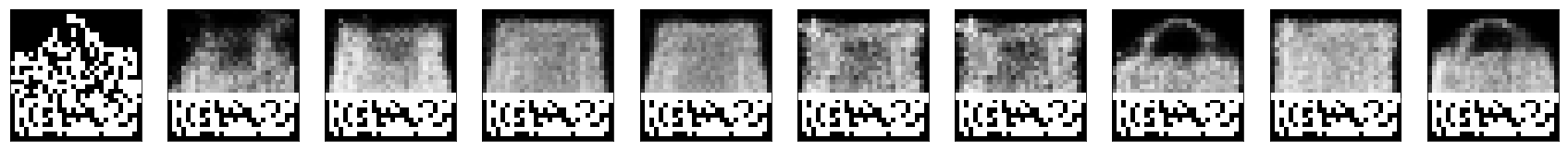}}\hfill%
	{\includegraphics[width=0.325\textwidth]{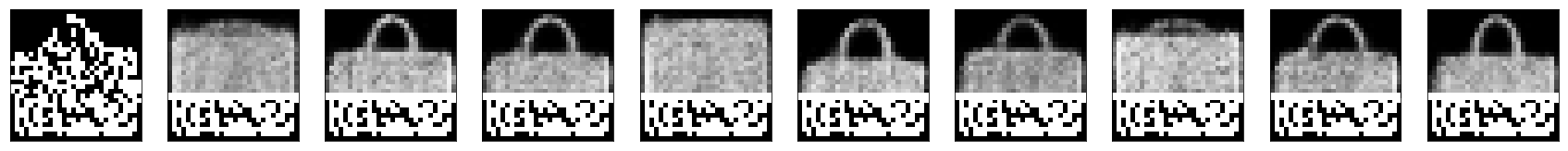}}\hfill%
	{\includegraphics[width=0.325\textwidth]{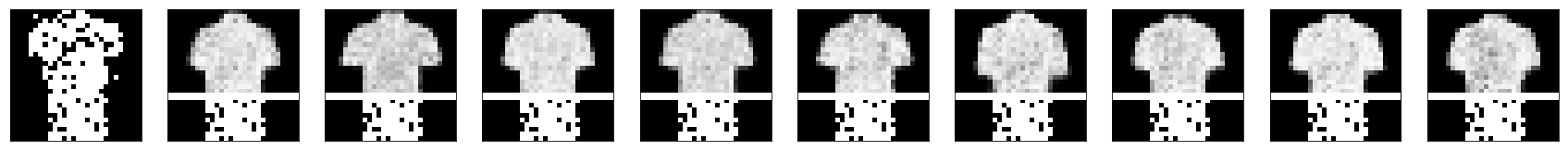}}\hfill%
	{\includegraphics[width=0.325\textwidth]{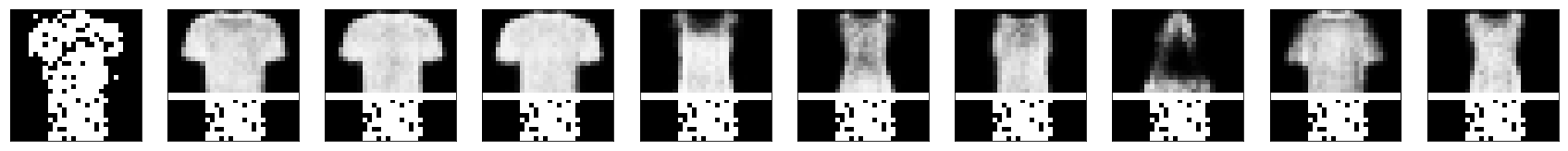}}\hfill%
	{\includegraphics[width=0.325\textwidth]{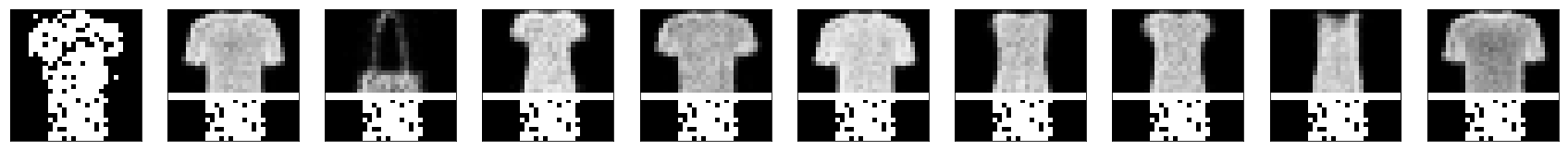}}\hfill%
	\vskip -11pt
	\subfloat[CVAE \label{fig:5-5a}] {\includegraphics[width=0.325\textwidth]{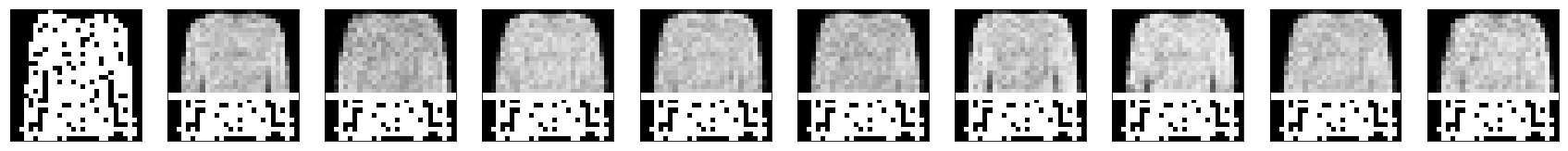}}\hfill%
	\subfloat[CMoG-CVAE \label{fig:5-5b}] {\includegraphics[width=0.325\textwidth]{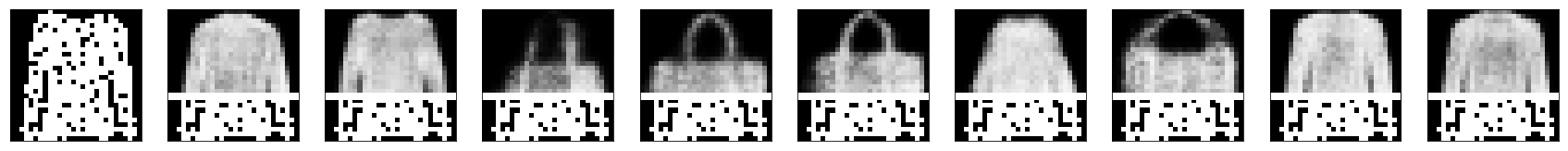}}\hfill%
	\subfloat[CDV-CVAE \label{fig:5-5c}] {\includegraphics[width=0.325\textwidth]{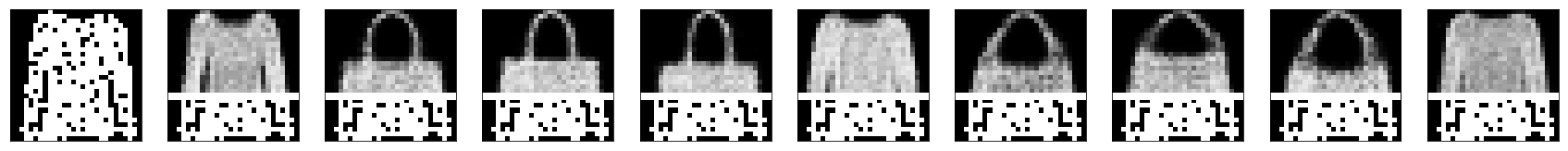}}\hfill%
	\caption{%
		Modified MNIST and Fashion-MNIST: %
		the goal is to validate whether the models can generalise and learn a one-to-many from a dataset of one-to-one mappings.
		The respective first column shows images of the test set, consisting of a condition~(lower third) and a target~(upper two-thirds).
		The remaining nine columns show generations conditioned on the lower third of the first image~(marked by the white line).
		The variety of generated targets in (b) and (c) is significantly larger than in (a).
		However, in case of the CMoG prior (b) we obtained a high amount of poor generations.
		(see Sec.~\ref{sec:experiments-gen})
	}%
	\label{fig:5-5}
\end{figure*}
\begin{figure*}[t]
	\centering	
	\subfloat[MNIST \label{fig:5-6a}] {\includegraphics[width=0.45\textwidth]{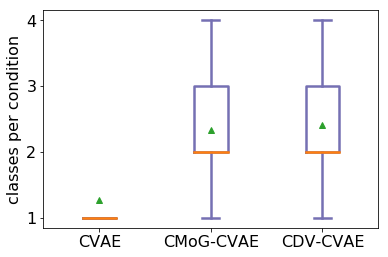}}\hfill%
	\subfloat[Fashion-MNIST \label{fig:5-6b}] {\includegraphics[width=0.45\textwidth]{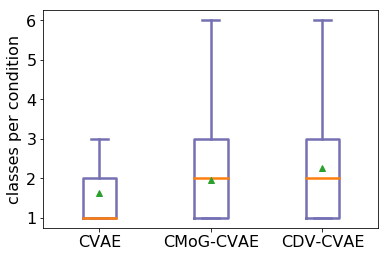}}\hfill%
	\caption{%
		Variety of generated targets: 
		for each condition in the test dataset, 10 targets were generated.
		A classifier was used to determine the number of different classes per condition.
		The box plots in (a) and (b) show that CMoG- and CDV-CVAEs generate targets with a larger variety for both datasets.
		(see Sec.~\ref{sec:experiments-gen})
	}%
	\label{fig:5-6}
\end{figure*}
\begin{figure*}[t]
	\centering	
	\subfloat[CMoG prior \label{fig:5-7a}] {\includegraphics[width=0.33\textwidth]{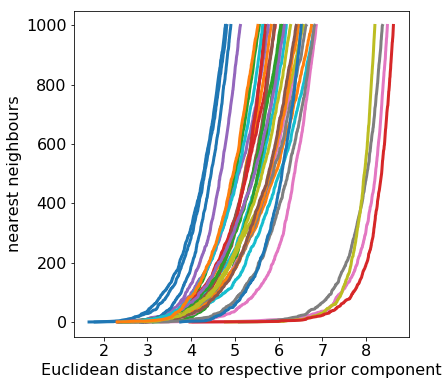}}\hfill%
	\subfloat[conditional VampPrior \label{fig:5-7b}] {\includegraphics[width=0.33\textwidth]{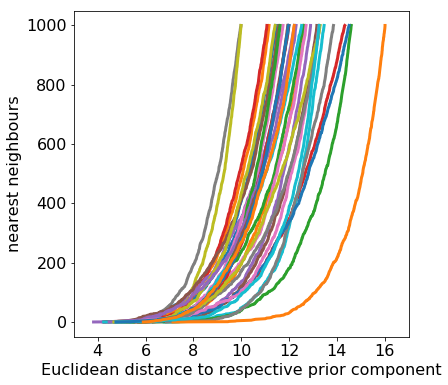}}\hfill%
	\subfloat[CDV prior \label{fig:5-7c}] {\includegraphics[width=0.33\textwidth]{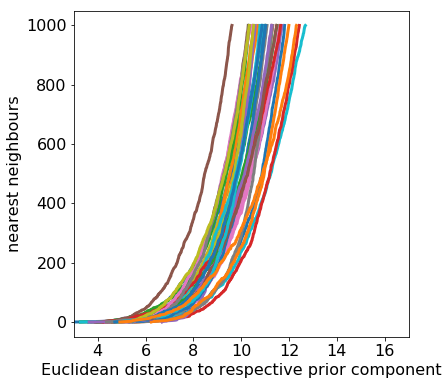}}\hfill%
	\caption{%
		The plots show the number of nearest neighbours~(encoded MNIST data points) as a function of the Euclidean distance to the mean of the respective mixture component.
		Each line belongs to one mixture component.
		Four mixture components of the CMoG prior~(a) and one of the conditional VampPrior~(b) have a significantly larger distance to the encoded data, reinforcing the conclusion that they focused on outliers during optimisation.
		Samples from one of these mixture components lead to poor generations like in Fig.~\ref{fig:5-5b}.
		(see Sec.~\ref{sec:experiments-gen})	
	}%
	\label{fig:5-7}
\end{figure*}
We created a modified version of MNIST~\cite{lecun1998gradient} and Fashion-MNIST~\cite{xiao2017fashion} to evaluate the generalisation capacity of the different models.
For this purpose, we split binarised MNIST/Fashion-MNIST images into two parts: a conditional part, the lower third~(last $28 \times 10$ pixels) of the image---and a target part, the upper two-thirds~(first $28 \times 18$ pixels).
The dataset has therefore only one target per condition.
The goal is to investigate whether the models are able to define a set of new targets for each condition of the test set.
In other words, whether they can learn a one-to-many from a one-to-one mapping.

In all three models, we used a 32-dimensional latent space.
CMoG-CVAEs (Fig.~\ref{fig:5-5b}) and CDV-CVAEs~(Fig.~\ref{fig:5-5c}) were able to represent a multimodal likelihood distribution, in contrast to vanilla CVAEs~(Fig.~\ref{fig:5-5a}).
This is shown by the significantly larger variety of generated targets per condition.

To measure the variety of the generated targets, we trained a classifier on MNIST/Fashion-MNIST and sampled 10 targets for each condition of the test set.
Afterwards, we used the classifier to determine how many different classes were generated per condition. 
Fig.~\ref{fig:5-6} shows the results for the different models and datasets.
Note that we only took sampled targets into account, which could be clearly assigned to a class---especially to avoid treating poor generations as additional classes.
In case of both datasets, CMoG- and CDV-CVAEs learned to generate several classes per condition, and thus a one-to-many from a one-to-one mapping.
Additionally, CMoG- and CDV-CVAEs achieved a larger variety of generations within the same class~(see Fig.~\ref{fig:5-5}).

Based on the above results, we can deduce that CMoG- and CDV-CVAEs have a higher generalisation capacity.
The larger variety of the generations is due to the structure of the priors:
since they are mixtures of $K$ distributions, each target is represented by one or more mixture components.
However, as discussed in Sec.~\ref{sec:methods-cdvp}, CMoG priors perform badly, especially in high dimensional latent spaces.
This becomes evident by the high amount of poor generations in Fig.~\ref{fig:5-5b}.
To verify our hypothesis---that the poor generations are caused by mixture components of the CMoG prior that focused on outliers during optimisation---we encoded our training data (MNIST) and measured the Euclidean distance to the respective mean of each prior component.
Fig.~\ref{fig:5-7} shows the number of nearest neighbours~(encoded data points) as a function of the Euclidean distance in the latent space.
Each line represents one of the 32 mixture components.
In contrast to the CDV prior~(Fig.~\ref{fig:5-7c}), four mixture components of the CMoG prior~(Fig.~\ref{fig:5-7a}) have a significantly larger distance to the encoded data.
This reinforces the conclusion that these mixture components focused on outliers during the optimisation process. 
We obtain poor generations like in Fig.~\ref{fig:5-5b} if a generated target is based on one of these four components, because $p_\theta(\x \givn \z)$ is only optimised (see Eq.~\ref{eq:elbo-cvae}) to decode samples that lie on the manifold of the encoded training data.

Additionally, we show that the CDV prior outperforms the conditional VampPrior~(Fig.~\ref{fig:5-7b}), where one mixture component has a significantly larger distance to the encoded data. 
As discussed in Sec.~\ref{sec:methods-cdvp}, we suspect this due to the higher dimension of the pseudo targets $\tilde{\mathbf{y}}$, making them more complex to optimise than pseudo latent variables $\tilde{\mathbf{z}}$.

\subsection{Generating Grasping Poses}
\label{sec:experiments-grasping}
\begin{figure*}[t]
	\centering	
	\subfloat[Test objects with proposed grasping poses \label{fig:5-9a}] {%
		\includegraphics[width=0.20\textwidth]{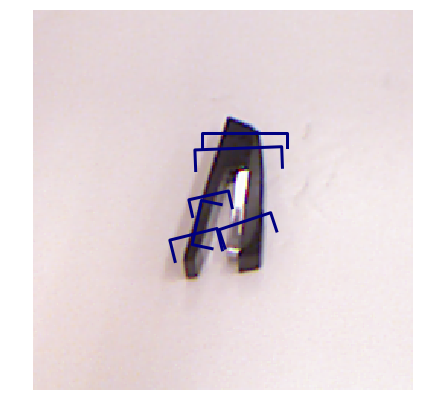}%
		\includegraphics[width=0.20\textwidth]{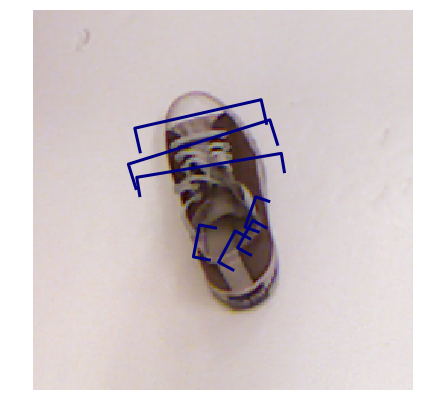}%
		\includegraphics[width=0.20\textwidth]{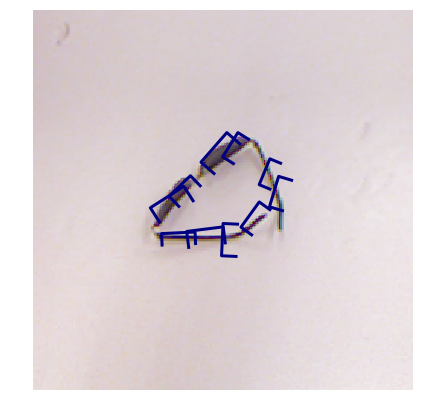}%
		\includegraphics[width=0.20\textwidth]{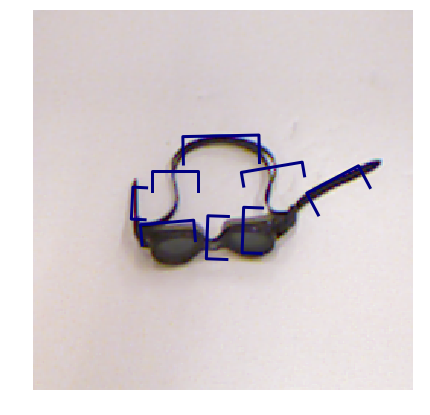}%
		\includegraphics[width=0.20\textwidth]{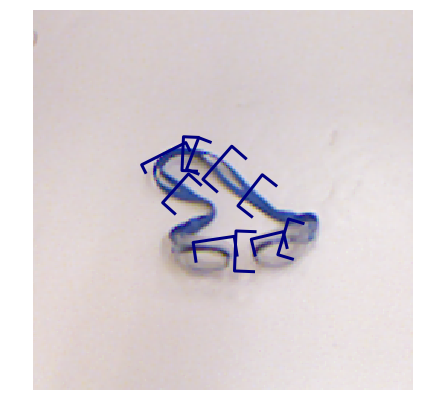}%
	}\hfill%
	\vskip 5pt	
	\subfloat[CVAE \label{fig:5-9b}] {%
		\includegraphics[width=0.20\textwidth]{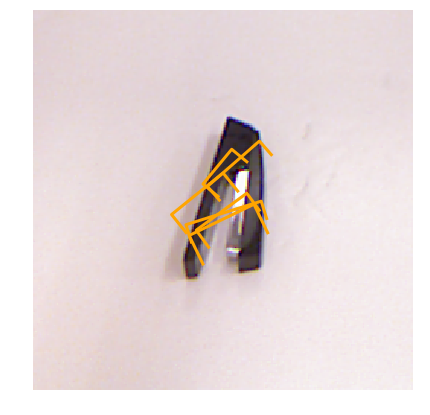}%
		\includegraphics[width=0.20\textwidth]{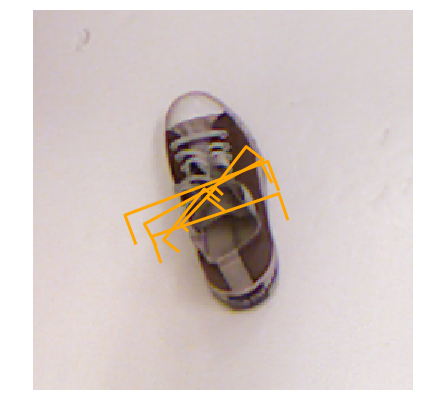}%
		\includegraphics[width=0.20\textwidth]{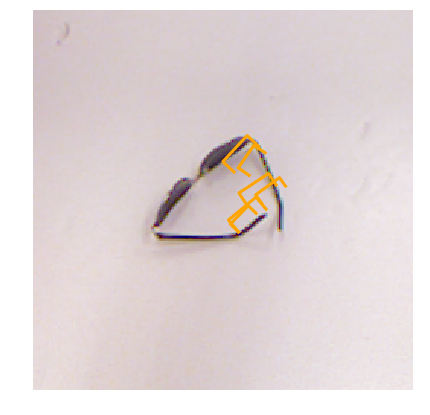}%
		\includegraphics[width=0.20\textwidth]{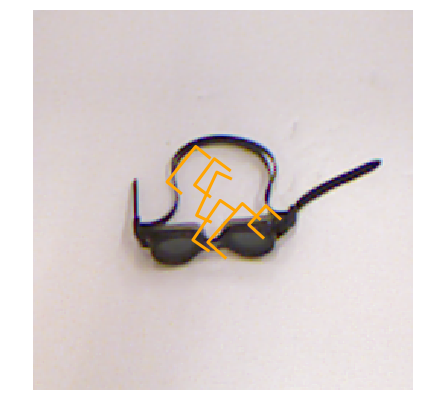}%
		\includegraphics[width=0.20\textwidth]{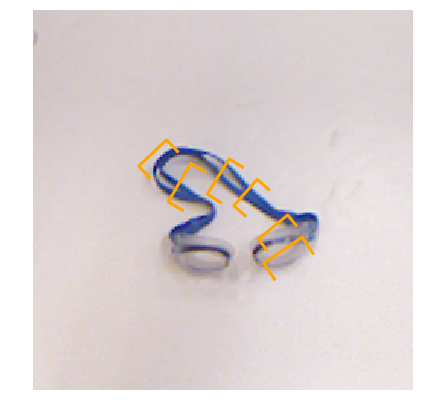}%
	}\hfill%
	\vskip 5pt			
	\subfloat[CDV-CVAE \label{fig:5-9c}] {%
		\includegraphics[width=0.20\textwidth]{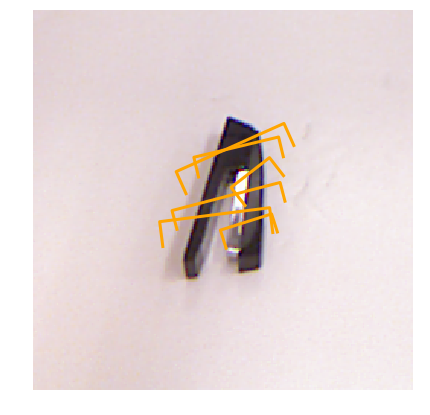}%
		\includegraphics[width=0.20\textwidth]{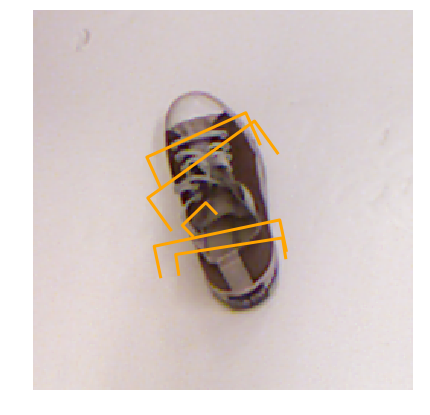}%
		\includegraphics[width=0.20\textwidth]{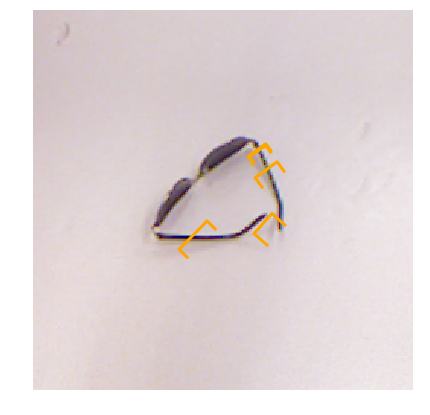}%
		\includegraphics[width=0.20\textwidth]{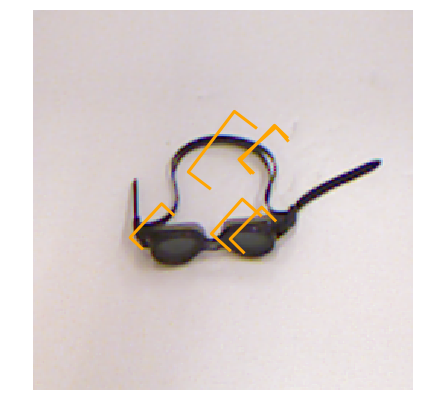}%
		\includegraphics[width=0.20\textwidth]{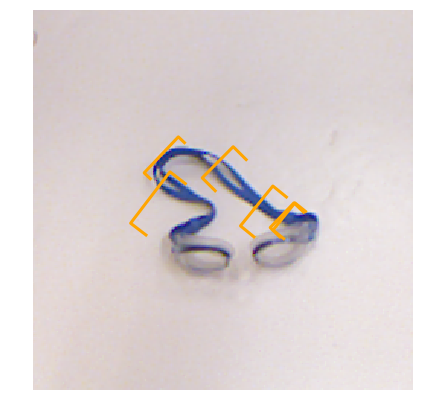}%
	}\hfill%
	\caption{%
		Cornell Robot Grasping dataset: 
		(a) objects (conditions) with proposed grasping poses (targets) defined by the test dataset.
		The CDV-CVAE~(c) generates more realistic grasping poses for unknown objects than the original CVAE~(a).
		$29\%$ of the grasping poses generated by the CDV-CVAE were above a discrimination score of 0.99, whereas the CVAE reached $22\%$.
		(see Sec.~\ref{sec:experiments-grasping})
	}%
	\label{fig:5-9}
\end{figure*}
In this experiment we want to assess the generalisation capabilities of CVAE and CDV-CVAE on a real-world dataset.
To this end, we use the Cornell Robot Grasping dataset, which consists of 885 conditions~($250 \times 250$ pixels greyscale images of objects) and 5,110 targets~(proposed grasping poses)~\cite{lenz2015}.
The latent spaces of both models are 16-dimensional.
For training, we resized the conditions to $64 \times 64$ pixels. 
Furthermore, we adapted the way how the grasping poses are represented:
the rectangles~(original representation) were redefined by a centre, a short and long axis, and a rotation angle.

Fig.~\ref{fig:5-9a} shows a selection of objects and proposed grasping poses defined by the test dataset.
Fig.~\ref{fig:5-9b} and Fig.~\ref{fig:5-9c} depict grasping poses generated by the CVAE and CDV-CVAE, respectively.
As discussed in Sec.~\ref{sec:experiments-zero} and \ref{sec:experiments-gen}, CDV-CVAEs have a higher capability of modelling one-to-many mappings and enable a larger variety of generated targets.

To verify whether the CDV-CVAE has actually learned to generate more realistic grasping poses for unknown objects, we apply a similar approach as proposed in~\cite{lenz2015}.
It is based on a discriminator for validating proposed grasping poses.
For this purpose, we trained the discriminator in equal parts with samples from joint and marginal empirical distribution $(\x, \y)\sim\hat{p}(\x, \y)$ and ${(\x, \y)\sim\hat{p}(\x)\,\hat{p}(\y)}$, respectively.
Subsequently, we generated 10 grasping poses for each condition in the test set and filtered out those with a discrimination score below 0.99.
As a result, $29\%$ of the grasping poses generated by the CDV-CVAE were above this threshold, whereas the CVAE reached $22\%$.
This allows the conclusion that the CDV-CVAE is a useful extension to the CVAE framework.

%% file: sections/conclusion.tex
\section{Conclusion}
\label{sec:conclusion}
In this paper, we have introduced a modified conditional latent variable model to incentivise informative latent representations.
To enable the model for capturing semantically meaningful features of the data, we have proposed an expressive multimodal prior that facilitates, in contrast to a classical Gaussian mixture prior, a well trained generative model. 

We have shown that our approach increases the generalisation capacity of CVAEs on a modified version of MNIST and Fashion-MNIST by achieving a significantly larger variety of generated targets---and on the Cornell Robot Grasping dataset by generating more realistic grasping poses.
Additionally, we have demonstrated that a straightforward application of CVAEs to structured-prediction problems suffers from a difficulty to represent multimodal distributions and that our approach overcomes this limitation.

%% file: MMP_CVAE.bbl
\begin{thebibliography}{10}
\providecommand{\url}[1]{\texttt{#1}}
\providecommand{\urlprefix}{URL }
\providecommand{\doi}[1]{https://doi.org/#1}

\bibitem{alemi2017fixing}
Alemi, A.A., Poole, B., Fischer, I., Dillon, J.V., Saurous, R.A., Murphy, K.:
  Fixing a broken {ELBO}. ICML  (2018)

\bibitem{bishop1997magnification}
Bishop, C.M., Svens'~en, M., Williams, C.K.I.: {Magnification factors for the
  SOM and GTM algorithms}. Proceedings Workshop on Self-Organizing Maps  (1997)

\bibitem{bowman2015generating}
Bowman, S.R., Vilnis, L., Vinyals, O., Dai, A.M., Jozefowicz, R., Bengio, S.:
  Generating sentences from a continuous space. CoNLL  (2016)

\bibitem{chen2016variational}
Chen, X., Kingma, D.P., Salimans, T., Duan, Y., Dhariwal, P., Schulman, J.,
  Sutskever, I., Abbeel, P.: Variational {Lossy Autoencoder}. CoRR  (2016)

\bibitem{higgins2017beta}
Higgins, I., Matthey, L., Pal, A., Burgess, C., Glorot, X., Botvinick, M.,
  Mohamed, S., Lerchner, A.: beta-{VAE}: Learning basic visual concepts with a
  constrained variational framework. ICLR  (2017)

\bibitem{Kingma2013}
Kingma, D.P., Welling, M.: Auto-encoding variational {B}ayes. CoRR  (2013)

\bibitem{krizhevsky2012imagenet}
Krizhevsky, A., Sutskever, I., Hinton, G.E.: Imagenet classification with deep
  convolutional neural networks. NeurIPS  (2012)

\bibitem{lecun1998gradient}
LeCun, Y., Bottou, L., Bengio, Y., Haffner, P., et~al.: Gradient-based learning
  applied to document recognition. Proceedings of the IEEE  (1998)

\bibitem{lenz2015}
Lenz, I., Lee, H., Saxena, A.: Deep learning for detecting robotic grasps. The
  International Journal of Robotics Research  (2015)

\bibitem{nalisnick2016stick}
Nalisnick, E., Smyth, P.: Stick-breaking variational autoencoders. ICLR  (2017)

\bibitem{pinto2016supersizing}
Pinto, L., Gupta, A.: Supersizing self-supervision: Learning to grasp from 50k
  tries and 700 robot hours. ICRA  (2016)

\bibitem{Rezende2014}
Rezende, D.J., Mohamed, S., Wierstra, D.: Stochastic backpropagation and
  approximate inference in deep generative models. ICML  (2014)

\bibitem{simonyan2014very}
Simonyan, K., Zisserman, A.: Very deep convolutional networks for large-scale
  image recognition. ICLR  (2015)

\bibitem{Sohn2015}
Sohn, K., Lee, H., Yan, X.: Learning structured output representation using
  deep conditional generative models. NeurIPS  (2015)

\bibitem{sonderby2016ladder}
S{\o}nderby, C.K., Raiko, T., Maal{\o}e, L., S{\o}nderby, S.K., Winther, O.:
  Ladder variational autoencoders. NeurIPS  (2016)

\bibitem{Tang2013}
Tang, Y., Salakhutdinov, R.R.: {Learning Stochastic Feedforward Neural
  Networks}. NeurIPS  (2013)

\bibitem{Tomczak2017}
Tomczak, J., Welling, M.: {VAE} with a {VampPrior}. AISTATS  (2018)

\bibitem{veres2017modeling}
Veres, M., Moussa, M., Taylor, G.W.: Modeling grasp motor imagery through deep
  conditional generative models. IEEE Robotics and Automation Letters  (2017)

\bibitem{walker2016uncertain}
Walker, J., Doersch, C., Gupta, A., Hebert, M.: An uncertain future:
  Forecasting from static images using variational autoencoders. ECCV  (2016)

\bibitem{xiao2017fashion}
Xiao, H., Rasul, K., Vollgraf, R.: Fashion-{MNIST}: a novel image dataset for
  benchmarking machine learning algorithms. arXiv:1708.07747  (2017)

\bibitem{yan2016attribute2image}
Yan, X., Yang, J., Sohn, K., Lee, H.: Attribute2image: Conditional image
  generation from visual attributes. ECCV  (2016)

\end{thebibliography}
